\documentclass[lettersize,onecolumn,journal]{IEEEtran}
\usepackage{amsmath,amsfonts}
\usepackage{algorithmic}
\usepackage{algorithm}
\usepackage{array}
\usepackage[caption=false,font=normalsize,labelfont=sf,textfont=sf]{subfig}
\usepackage{textcomp}
\usepackage{stfloats}
\usepackage{url}
\usepackage{verbatim}
\usepackage{graphicx}
\usepackage{cite}
\usepackage{multirow}
\usepackage{tabularx}
\usepackage{xcolor}

\newcommand{\tabincell}[2]{\begin{tabular}{@{}#1@{}}#2\end{tabular}}

\usepackage{hyperref}
\usepackage[nameinlink]{cleveref}
\Crefname{figure}{{Fig.}}{{Figs.}}
\Crefname{table}{{Table.}}{{Tables.}}
\Crefname{equation}{{Eq.}}{{Eqs.}}
\Crefname{Algorithm}{}{}

\begin{document}

\title{Exploring the Generalizability of Geomagnetic Navigation: A Deep Reinforcement Learning approach with Policy Distillation}

\author{Wenqi Bai, Shiliang Zhang,~\IEEEmembership{Member,~IEEE}, Xiaohui Zhang, Xuehui Ma, Songnan Yang,~\IEEEmembership{Student Member,~IEEE}, Yushuai Li, \IEEEmembership{Senior Member, IEEE}, Tingwen Huang, \IEEEmembership{Fellow, IEEE}
\thanks{W. Bai, X. Zhang, X. Ma, and S. Yang are with Xi’an University of Technology, Xi'an, China (email: bayouenqy@outlook.com, xhzhang@xaut.edu.cn, xuehui.yx@gmail.com, yang.son.nan@gmail.com). S. Zhang is with University of Oslo, Oslo, Norway (e-mail: shilianz@ifi.uio.no). Y. Li is with Aalborg University, Aalborg, Denmark (yushuaili@ieee.org). T. Huang is with Shenzhen University of Advanced Technology, Shenzhen, China (e-mail: huangtingwen2013@gmail.com)}}



\maketitle

\begin{abstract}
The advancement in autonomous vehicles has empowered navigation and exploration in unknown environments. Geomagnetic navigation for autonomous vehicles has drawn increasing attention with its independence from GPS or inertial navigation devices. 
While geomagnetic navigation approaches have been extensively investigated, the generalizability of learned geomagnetic navigation strategies remains unexplored. The performance of a learned strategy can degrade outside of its source domain where the strategy is learned, due to a lack of knowledge about the geomagnetic characteristics in newly entered areas. This paper explores the generalization of learned geomagnetic navigation strategies via deep reinforcement learning (DRL). Particularly, we employ DRL agents to learn multiple teacher models from distributed domains that represent dispersed navigation strategies, and amalgamate the teacher models for generalizability across navigation areas. We design a reward shaping mechanism in training teacher models where we integrate both potential-based and intrinsic-motivated rewards. The designed reward shaping can enhance the exploration efficiency of the DRL agent and improve the representation of the teacher models. Upon the gained teacher models, we employ multi-teacher policy distillation to merge the policies learned by individual teachers, leading to a navigation strategy with generalizability across navigation domains. We conduct numerical simulations, and the results demonstrate an effective transfer of the learned DRL model from a source domain to new navigation areas. Compared to existing evolutionary-based geomagnetic navigation methods, our approach provides superior performance in terms of navigation length, duration, heading deviation, and success rate in cross-domain navigation.

\end{abstract}


\section{Introduction}

Geomagnetic navigation leverages the ubiquitous earth magnetic field signals for the navigation~\cite{zhao2014long,DBLP:journals/corr/abs-2412-11882}, without independence on dedicated devices along the navigation route~\cite{qi2023geographic,chen2023geomagnetic,zhang2020geomagnetic}. Geomagnetic navigation thus can secure the navigation mission, \textit{e.g.}, in remote areas or underwater environments where there GPS or pre-deployed navigation devices is unavailable~\cite{canciani2021magnetic}. With less dependence on external devices, geomagnetic navigation approaches are more self-sufficient and less susceptible to outer interference~\cite{karshakov2020promising}. Such advantages make geomagnetic navigation a promising solution especially in long-range missions into unexplored areas~\cite{DBLP:journals/corr/abs-2403-08808}.

Despite the advantages, it is challenging to maintain the generalizability of a learned geomagnetic navigation strategy. The challenge lies in that the carrier need to exploit the navigation strategy to approach their destination while simultaneously exploring the geomagnetic field in optimizing their navigation actions~\cite{zhang2020geomagnetic}. Such a navigation mission can be constrained by the absence of prior geomagnetic maps in unexplored environments~\cite{zhao2014long}. Furthermore, the geomagnetic field is influenced by various factors of uncertainties~\cite{DBLP:journals/tac/MaZLQSH24,MA2022157,9189668,DBLP:conf/eucc/MaCZLQS24} and even outliers~\cite{DBLP:journals/cssc/ZhangCYZH19,DBLP:journals/tnn/ZhangCYZH18}, \textit{e.g.}, the Earth's interior composition and rotation that can result in irregular geomagnetic patterns~\cite{xu2003natural}. This variability means that a well learned navigation strategy may struggle to adapt to new geomagnetic environments, risking failures in navigation missions.

The configuration of a learned navigation strategy should be generalized rather than dedicated to a narrow area with specific geomagnetic features. The generalizability of geomagnetic navigation approaches is critical to guarantee that the navigation strategy works beyond the area where the strategy is derived, thus avoiding case-wise and frequent model training~\cite{zhu2021deep}. In the following we review the generalizability of existing geomagnetic navigation approaches, and provide insights into the challenges yet to be addressed. 

Geomagnetic matching approaches determine the carrier location by comparing the magnetic field around the carrier with a pre-established geomagnetic map, using techniques \textit{e.g.}, mean absolute difference (MAD)~\cite{jia2012simulation} and mean squared deviation (MSD)~\cite{xie2013fast}, and contour matching~\cite{chen2018new,xu2022innovative}. While geomagnetic matching offers highly accurate carrier location for the navigation~\cite{karshakov2020promising}, they require the storage of the detailed and complete geomagnetic information within the navigation area that supports no generalizability. \textit{I.e.}, the navigation will fail once the carrier moves beyond the boundaries of the geomagnetic map. 


Bionic geomagnetic navigation methods draw inspiration from animal navigation~\cite{wiltschko2007magnetoreception,lohmann1996orientation,putman2015inherited,boles2003true}, where the animals navigate using the Earth geomagnetic field without a geomagnetic map stored in their brains. Such approaches generally navigate through trial-and-error where they adjust the navigation based on the feedback on their trial actions. Early bionic geomagnetic navigation methods develop heuristic algorithms to for path search. Liu \textit{et al.} formulated bionic geomagnetic navigation as an autonomous path search in a geomagnetic environment, and they developed a multi-objective search approach to identify optimal solutions \cite{liu2013bio}. Their simulation results demonstrate applicability of their approach in navigation, yet the resulted navigation trajectories exhibit zigzag patterns due to inefficient path search. Li \textit{et al.} proposed a path search based on evolutionary strategies~\cite{li2017bio} and reduced chattering path in the navigation, while their navigation routes are not smooth due to the randomness of the search strategy. Zhou \textit{et al.} proposed a differential evolution approach to enhance search efficiency \cite{zhou2022bionic} by refining the mutation mechanism in evolutionary algorithm, leading to reduced traveled distance in the navigation. Bionic geomagnetic navigation avoids pre-stored maps dedicated to specific navigation areas and is with improved generalizability. However, such bionic approaches depend on stochastic search that renders erratic trajectories. Bionic geomagnetic navigation is also sensitive to the path search configuration that, a premature convergence of the search may lead to suboptimal and repeated navigation actions by the carrier, resulting in detours or even mission failures.

Reinforcement learning based geomagnetic navigation has gain momentum in recent years in addressing unmodeled uncertainties in geomagnetic environments~\cite{10713176}. In contrast to geomagnetic match or bionic approaches that either relies on fixed prior knowledge or path search, reinforcement learning approaches aim to learn a navigation model that can be re-used in explored navigation areas. Wang \textit{et al.} proposed a geomagnetic navigation based on deep reinforcement learning~\cite{wang2019geomagnetic} that explores the correlation between geomagnetic parameters and navigation trajectories. They employed deep Q-network~\cite{10055961} that takes input as the geomagnetic observations and predicts heading angles for the navigation. They extended their work by introducing a deep double-Q-network framework~\cite{wang2020geomagnetic} with improved model training efficiency. However, the state representation in their methods only considers the geomagnetic observations around the vehicle's position, without incorporating observations around the destination. This design inherently limits their approach to a specific destination, jeopardizing its generalizability and demanding model re-training for new tasks. Bai \textit{et al.} adopted goal-oriented reinforcement learning where they develop trajectory exploration targeting generalized navigation areas, rather than fixed start-to-target trajectory planning~\cite{Bai2024LongdistanceGN}. 
Nevertheless, their approach merely optimizes an learning agent's actions within specific geomagnetic environments and attributes the model parameters to a narrow range of geomagnetic features. 
Li \textit{et al.} developed a deep Q-network approach for geomagnetic navigation with enhanced state representation and exploration by the learning agent~\cite{hong2024research}, leading to improved adaptability and robustness of the navigation in unfamiliar areas. However, their approach relies on accurate localization that necessitates additional positioning technologies, which contradicts the principle of geomagnetic navigation that avoids external aids. 
Overall, while the learned models by the previous reinforcement learning approaches can represent the gained knowledge for the navigation, it remains unexplored that to what level the gained knowledge is generalized and can be useful in new navigation missions. There is no mechanism that analyzes the generalizability of a trained inforcement learning model for geomagnetic navigation.

To narrow the gap, this paper explores the generalizability of the geomagnetic navigation by developing a deep reinforcement learning-based navigation. We design the approach to achieve generalizability of the geomagnetic navigation cross navigation areas. In the developed approach, we adopt policy distillation to emerge multiple learning agents trained in different navigation areas, facilitating model compression and robust model representativeness. The distilled knowledge represented by the merged model can be transferred to unexplored navigation areas, leading to enhanced model adaptability and generalizability. In this way, we empower the deep inforcement learning for geomagnetic navigation where we learn from extensive training data and transfer the learned knowledge to more generalized forms. We summarize our contributions in this work as follows.

\begin{itemize}
    \item We are the first to examine the generalizability of a trained geomagnetic navigation model across navigation areas. We explore the generalizabiltiy of the geomagnetic navigation by developing a deep reinforcement learning (DRL) approach to represent the learned knowledge for geomagnetic navigation.
    \item We train multiple DRL agents - or teacher models - in dispersed navigation areas, and we adopt policy distillation to corporate the teacher models and transfer the learned knowledge to a generalized form for geomagnetic navigation.
    \item We design the DRL agents with a reward shaping mechanism where we integrate both potential-based and intrinsic-motivated rewards. In this way, we facilitate the DRL agents with enhanced exploration efficiency and representativeness during the teacher model training, leading to learned teacher models with adaptability across navigation areas.
        
\end{itemize}

The rest of the article is organized as follows. Section~\ref{Section:2} presents the mathematical description of the geomagnetic navigation from a deep reinforcement learning perspective. Section~\ref{Section:3} details the proposed DRL approach for the generalizability of the geomagnetic navigation, and illustrates policy distillation for the DRL agents and the designed reward shaping mechanism. Section~\ref{Section:4} validates the effectiveness and accuracy of the proposed approach through simulations. Finally, Section~\ref{Section:5} summarizes findings from the simulation results and discusses future research.

\section{Fundamentals for Geomagnetic Navigation}\label{Section:2}

\subsection{Mathematical Description of the geomagnetic field}
\label{Section:2.1}
The geomagnetic field (GF) represents the magnetic structure of the Earth, resulting from the combination of magnetic fields generated by rocks within the Earth and electric currents distributed both internally and externally. It is primarily a dipole magnetic field distributed across planetary scales, it theoretically corresponds to the geographical position of any point in near-Earth space \cite{tyren1982magnetic}. The strength of the geomagnetic field is typically measured in nanoteslas (\(nT\)). 

\begin{figure}[h]
    \centering
    \includegraphics[width=0.75\linewidth]{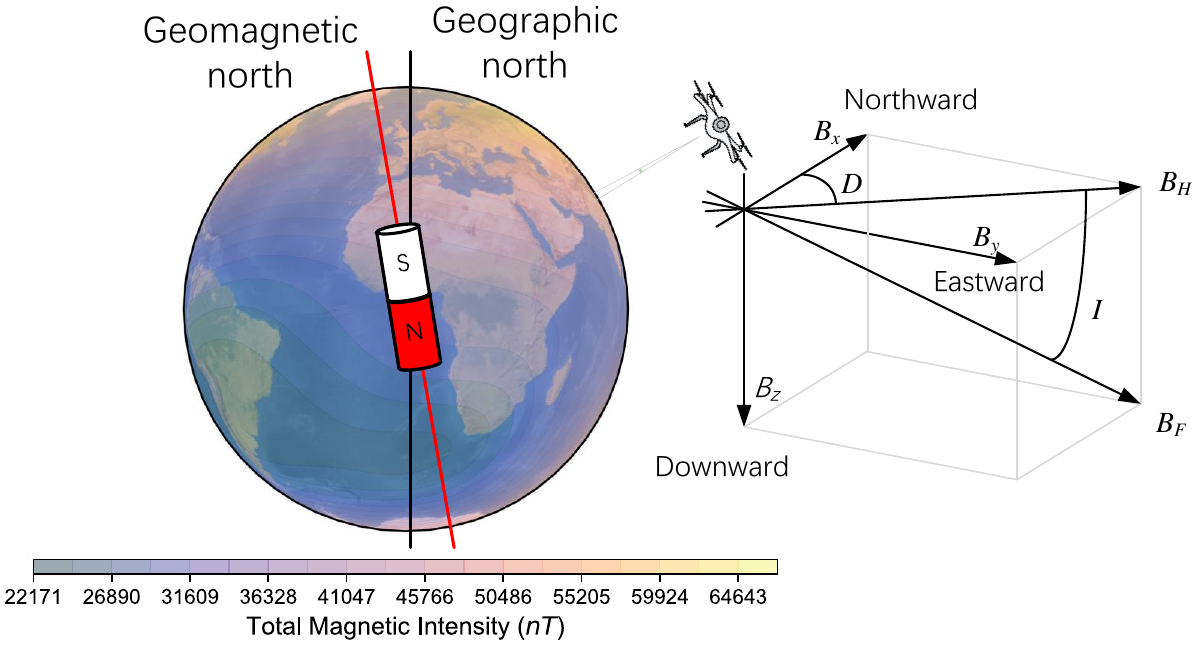}
    \caption{Description of the geomagnetic field and geomagnetic parameters.}
    \label{fig:geomagnetic field}
\end{figure}

The Geomagnetic North Pole, located on the northern hemisphere surface of the Earth, is where the geomagnetic field vector points vertically downward \cite{dawson1982magnetic}, deviating southwestward from the Geographic North Pole. The axes of the geomagnetic dipole and the Earth’s rotation form an 11-degree angle, as depicted in \Cref{fig:geomagnetic field}, where the Earth is represented by a spherical coordinate system, with its zenith axis directed toward the Geographic North Pole. The total intensity of the GF at each point can be denoted as \({B_F}\), which can be further decomposed into three orthogonal components. The north geomagnetic field component \({B_x}\), the east geomagnetic field component \({B_y}\), and the vertical geomagnetic field component \({B_z}\), pointing to the geographic east, the north, and the core of the Earth, respectively, the horizontal component of the GF is denoted as \({B_H}\), it can be decomposed into \({B_x}\) and \({B_y}\) in the horizontal plane. Furthermore, the included angle between the geographic north and \({B_H}\) is defined as the geomagnetic declination \({D}\), and the included angle between the horizontal plane and the BF is defined as the geomagnetic inclination \({I}\). Given any three of the seven geomagnetic parameters mentioned above, the remaining elements can be calculated by the following equations:
\begin{align}
\left\{\begin{array}{l}
B_x=B_F \cos I \cos D \\
B_y=B_F \cos I \sin D \\
B_z=B_F \sin I \\
B_F=\sqrt{B_x^2+B_y^2+B_z^2} \\
B_H=B_F \cos I \\
I=\arccos \left(B_H / B_F\right) \\
D=\arccos \left(B_x / B_H\right) .
\end{array}\right.
\label{eq:2.1}
\end{align}
On a macroscopic scale, the intensity of the geomagnetic field increases with latitude, reaching its lowest point near the equator at approximately 25,000\(nT\) and its highest point at the poles, approaching 65,000\(nT\), with a variation of around 3 to 5\(nT\) per kilometer. Similarly, the magnetic inclination angle increases with latitude, being close to 0 near the equator and reaching 90 and -90 degrees at the North and South Poles, respectively, with a variation of about 0.01 degrees per kilometer. In regions of lower latitudes, the magnetic declination is less pronounced, allowing the magnetic field direction to roughly align with the north-south axis. These distribution characteristics serve as the basis for implementing geomagnetic navigation.

\subsection{Geomagnetic Navigation}
\label{Section:2.2}
Geomagnetic navigation can be described as the prediction of a sequence of navigation actions taking input geomagnetic information. By taking in the geomagnetic field signals around the location of the autonomous vehicle and comparing them with the known geomagnetic field parameters at the destination, along with historical navigation state and action sequences, the current navigation action is predicted to guide the vehicle to the destination. As autonomous vehicles navigate through large-scale spaces, we simplify their representation to particles and focus on two-dimensional Cartesian coordinate systems for navigation problems. The movement of the vehicle is expressed as
\begin{align}
    \left\{ \begin{array}{l}
    {x_{j + 1}} = {x_{j}} + {L_{j+1}}\cos ({\phi _{j+1}})\\
    {y_{j + 1}} = {y_{j}} + {L_{j+1}}\sin ({\phi _{j+1}})
    \end{array} \right.
\label{eq:2-2}
\end{align}
where \({x_j}\), \({y_j}\) denote the position at time step \(j\), \({L_j}\) and \({\phi _j}\) respectively represent the movement distance and angular displacement of the vehicle in the Cartesian coordinate system. 

In this article, the angular displacement of the vehicle \({\phi _j}\) in the Cartesian coordinate system is estimated based on the yaw angles \({\psi _j}\) of the vehicle at adjacent time steps. Therefore, the movement equation is represented as: 
\begin{align}
    \left\{ \begin{array}{l}
    {x_{j + 1}} = {x_j} + {L_{j+1}}\cos ({\phi _j+\psi _{j+1}})\\
    {y_{j + 1}} = {y_j} + {L_{j+1}}\sin ({\phi _j+\psi _{j+1}})
    \end{array} \right.
\label{eq:2-3}
\end{align}
In the geomagnetic navigation task, the agent takes the yaw angles \({\psi _j}\) and the movement distance \({L_j}\) as the navigation action, which is determined by the agent based on one or more of  the geomagnetic parameters, as shown in   \Cref{eq:2-1}  at its current position and destination serve as state information, which can be denoted by an \(i\)-dimensional vector:
\begin{align}
\mathbf{B} = \{ {B_1},{B_2}, \ldots ,{B_i}\}
\label{eq:2-4}
\end{align}
The agent iteratively selects navigation actions including the yaw angle \({\psi _j}\) and the movement velocity  \({v_j}\) to maximize rewards, guiding the vehicle to converge from the value of the current position to the destination. When the geomagnetic multi-parameter converges to zero, it signifies that the vehicle has reached the destination and completed the navigation task\cite{liu2013bio}, that is:
\begin{align}
    \lim _{k \rightarrow \infty} F(\mathbf{B}, k) \rightarrow 0
    \label{eq:2-5}
\end{align}
where \(F(\mathbf{B}, j)\) is  the normalized objective function at the \(j\)-th time step, which can be expressed as:
\begin{align}
F(\mathbf{B}, j)=\sum_{i=1}^n \frac{f_i^j(\mathbf{B}, j)}{f_i^{\prime}(\mathbf{B}, j)}=\sum_{i=1}^n \frac{\left(\mathbf{B}_i^{T}-\mathbf{B}_i^j\right)^2}{\left(\mathbf{B}_i^{T}-\mathbf{B}_i^0\right)^2}
\label{eq:2-6}
\end{align}  
\(\mathbf{B}_i^j\) and  \(\mathbf{B}_i^{T}\) are the value of the \(i\)-th geomagnetic parameter at the current position and destination, respectively. \(f_i(\mathbf{B}, j)\)  represents the objective sub-function of the \(i\)-th geomagnetic parameter at the \(j\)-th time step. 

In actual navigation, when the distance between the vehicle and the destination is within a certain range, the vehicle is considered to navigate to the destination successfully. This can be equivalent to the total objective function being less than the set threshold \(\zeta\). 
\begin{align}
    F(\mathbf{B}, j)<\zeta
\label{eq:2-7}
\end{align}

\subsection{Formulation of DRL for Geomagnetic Navigation}
\label{Section:2.3}

From the perspective of DRL, a robust and constrained association exists between geomagnetic parameters and the navigation path. Consequently, the DRL facilitates the training of the agent to acquire a "magnetotactic" capability akin to that observed in migrating animals. However, our previous research has shown that the navigation strategy learned by the agent in a specific region is only effective within that region and deteriorates significantly in scenarios outside the training region due to overfitting of the agent to the training environment. Therefore, the agent should adopt an agent-centric representation of behavior with consistent semantics across regions, rather than an environment-centric one whose semantics may change across regions. This means that the model should establish a convergence relationship for geomagnetic parameters throughout the navigation by leveraging the sensitivity of geomagnetic trends to these elements.

To illustrate the impact of deep reinforcement learning-based geomagnetic navigation models on navigation performance in unexplored regions, we conducted two experiments on a geomagnetic navigation environment built on Gym using the DQN implemented in OpenAI Baselines3.  We selected a tropical ocean region in the Atlantic Ocean, spanning from -43 to -33 degrees longitude and 10 to 20 degrees latitude. The agent was trained in this region, and we evaluated the navigation performance of both the trained and untrained agents in the training and unknown regions. The replay buffer size was set to 50,000, and the agent was trained for 10k environment steps. The results indicated that while the agent could quickly reach the destination in the training region, its navigation performance significantly declined in the unknown region, with a high risk of failures. 

Given this, it's necessary to extend agents trained for specific navigation tasks to more general and complex navigation regions. We consider a family of Markov decision processes (MDP) \(\mathcal{M} =\left \{ M_i \right \}^N_{i=1}\) consisting of \(N\) agents trained from different regions. Each MDP can be denoted by \(M_i=\left \langle \mathcal{S_i},\mathcal{A_i},\mathcal{T_i},r_i,\gamma,\rho _0 \right \rangle\), where \(\mathcal{S_i}\) and \(\mathcal{A_i}\) denote the source state space and source action space for the \(i\)-th agent, \(\gamma\) and \(\rho _0\) represent the discount factor and the distribution of initial states, \(\mathcal{T_i}:\mathcal{S_i}\times \mathcal{A_i}\times \mathcal{S_i}\to [0,1] \) is the dynamics transition function from current state \(s\) to next state \(s'\) under action \(a\), \(\mathcal{r_i}:\mathcal{S_i}\times \mathcal{A_i}\to \mathbb{R} \) is the reward function. Considering the generalization of navigation tasks, agents need to learn common knowledge from policies in different regions and develop a unified policy to reach them. The formulation of goal-conditioned reinforcement learning (GCRL) is introduced to augment the Markov Decision Process (MDP) with an extra tuple \(\left \langle \mathcal{G},p_g,\varphi \right \rangle\), forming a goal-augmented MDP (GA-MDP)\cite{plappert2018multi}. In this formulation, \(\mathcal{G}\) represents the goals space of navigation tasks, \(p_g\) denotes the desired goal distribution of the environment, and \(\varphi:\mathcal{S} \to \mathcal{G}\)  is a function that maps states to specific goals, thus the objective is to learn a policy \(\pi_\theta:\mathcal{S}\times \mathcal{G}\times \mathcal{A}\to [0,1]\) with internal parameter \(\theta \in \Theta\) that maximizes the expectation of the cumulative return over the goal distribution:
\begin{align}
J(\pi)=\mathbb{E}_{a_t\sim \pi_\theta (\cdot |s_t,g),g\sim p_g,s_{t'}\sim\mathcal{T}(\cdot |s_t,a_t)}\left [ \sum_{t}^{} \gamma ^tr(s_t,a_t,g) \right ]
\label{eq:2-6}
\end{align}
This policy aims to acquire a transferable strategy so that the agent can navigate even in unknown regions, similar to migratory animals, which can eventually reach their destination even in unexplored territories.

\section{Domain Generalization by DRL with Policy Distillation}\label{Section:3}
In this section, we provide a detailed description of the proposed DRL approach 
designed to train a geomagnetic navigation model with generalization capabilities. We adopt the Twin Delayed Deep Deterministic Policy Gradient (TD3) algorithm for the DRL agent training, and we then employ policy distillation and derive our approach of TD3 with separate training and ensemble policy distillation (TD3-STEPD) to merge multiple trained DRL agents for enhanced generalizability. First, we describe the overview of the proposed algorithm, which comprises two components: multi-localized navigation networks and a global navigation network. Then, we detail the implementation of the multi-teacher policy distillation employed in training the navigation model, along with measures to improve the efficiency of DRL agent exploration.

\subsection{Overall Framework of TD3-STEPD}

\begin{figure}[h]
    \centering
    \includegraphics[width=0.7\linewidth]{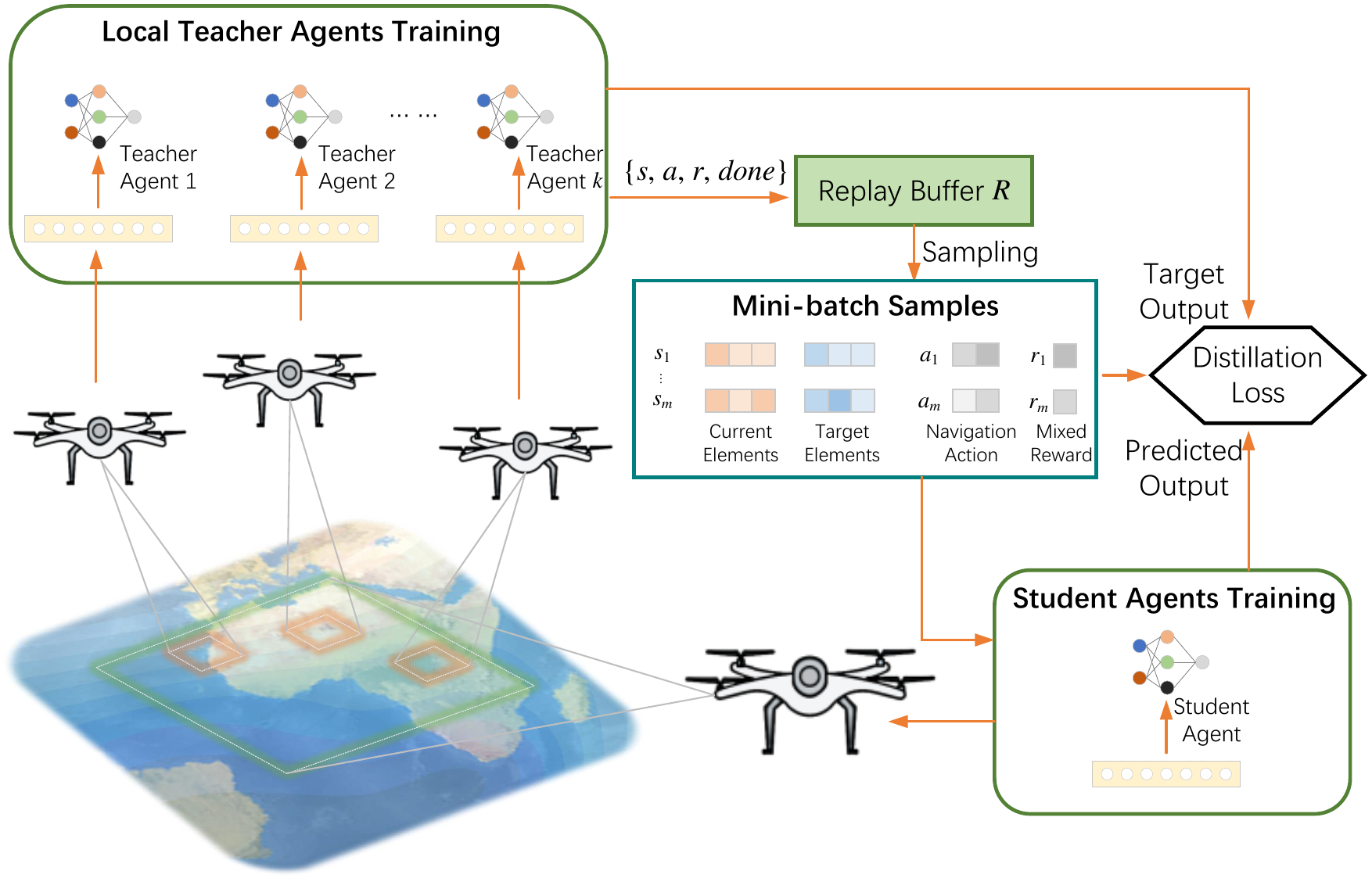}
    \caption{This framework consists of teacher networks and a student network. The top part demonstrates the \textit{k} teacher policy networks, independently trained in different regions. The orange boxes on the map indicate the training regions of the \textit{k} teacher policy networks. The bottom part shows how policy distillation integrates the navigation strategies from the \textit{k} teacher policy networks into the student policy network, thereby extending the navigation domain to a larger region as indicated by the green box on the map.}
    \label{fig:framework}
\end{figure}

We illustrate the structure of our TD3-STEPD algorithm in \Cref{fig:framework}. It consists of multiple teacher networks and a student network. Considering that both the state space and action space are continuous in the geomagnetic navigation task, the entire network is trained using the Twin Delayed Deep Deterministic Policy Gradient (TD3) algorithm. TD3 is a robust DRL algorithm \cite{fujimoto2018addressing}, combining the advantages of DQN and the actor-critic (AC) structure, proving more stable than DDPG. The fundamental relationship between the value of a state-action pair \((s_t,a_t)\) and the subsequent state-action pair \((s_{t+1},a_{t+1})\) in Q-learning is described by the Bellman equation:
\begin{align}
Q^\pi(s_t,a_t)=r(s_t,a_t)+\gamma \mathbb{E}_{s_{t+1},a_{t+1}}[Q^\pi(s_{t+1},a_{t+1})],\\ \quad a_{t+1}\gets \pi (s_{t+1})
\label{eq:2-7}
\end{align}
The TD3 algorithm creates two differentiable function approximators to estimate the value function \(Q^ \pi\), during parameter optimization, it selects the minimum estimate to address the issue of overestimation of the \(Q^ \pi\), when the target network \(Q_ {\phi_i}'\) is updated of Clipped Double Q-learning algorithm \cite{fujimoto2018addressing}:
\begin{align}
y=r+\gamma \min\limits _{i=1,2}Q_{\phi '_i}(s_{t+1},\pi_{\theta}(s_{t+1}))
\label{eq:2-7}
\end{align}
let \(\phi_i'\) and \(\phi_i\) define the parameters of the target critic network \(Q_{\phi '_i}\) and training critic network \(Q_ {\phi_i}\), respectively. The \(\phi_i\) is updated by randomly sampling batches from replay memory and regressing to this target:
\begin{align}
    \mathcal{L}_{\mathcal{D},\phi_i}=\mathop{E}\limits_{e\sim \mathcal{D}}[(Q_{\phi_i}(s_t, a_t)-y)^2]
\label{eq:2-7}
\end{align}
and the parameters of the policy network \(\theta\) are learned just by the deterministic policy gradient algorithm\cite{silver2014deterministic} ascent to maximize \(Q_{\phi_1}\):
\begin{align}
    \max\limits_\theta \mathop{E}\limits_{e\sim \mathcal{D}}[Q_{\phi_1}(s_t, \pi_\theta(s))]
\label{eq:2-7}
\end{align}

We adopt \(k\) TD3 policy network architectures as our  teacher networks denote as \(T_k\), each policy network utilizes MLP to capture meaningful attributes from the input vector and determine the navigation action for the next time step, learning the localized navigation strategy within their respective navigation domains. Subsequently, the student network \(S\) leverages the acquired strategies from multiple teacher networks, as detailed in the next subsection, employing policy distillation to train a generalized navigation strategy insensitive to navigation regions. This strategy is about to address magnetic navigation tasks for autonomous mobile vehicles in unexplored regions. 

We selected three geomagnetic parameters: geomagnetic declination \({D}\), geomagnetic inclination \(I\), and horizontal component \({B_H}\) from the seven parameters shown in \Cref{eq:2.1}, which can be observed by the magnetometer carried on autonomous mobile vehicles. Compared to standard RL solutions that learn a policy based solely on states or observations, geomagnetic navigation tasks additionally require the agent to make decisions based on varying goals, which can be seen as goal-conditioned RL\cite{plappert2018multi}. Thus, the geomagnetic parameters of the destination are augmented to the state (i.e., the geomagnetic parameters measured at the current position) for decision-making, and the input vector is expressed as \(I^j=[D^j,I^j,B_H^j,D^d,I^d,B_H^d]\). We utilize MLP to capture meaningful attributes from the input vector and determine the navigation action for the next time step. It is worth noting that the proposed algorithm is not tied to TD3 and can be applied to other off-policy reinforcement learning frameworks.

\subsection{Policy distillation}
Policy distillation has been proposed to compress network size without performance degradation, combine multiple expert policies into a single multi-task policy surpassing individual experts, and enable real-time, online learning by continuously distilling the best policy to a target network~\cite{rusu2015policy}. In this paper, we consolidate actor policies by devising a distilled policy aimed at minimizing its Kullback-Leibler (KL) divergence across all agents. We amalgamate observation and action spaces from all agents and treat the geomagnetic navigation problem as a unified, higher-dimensional Markov decision process\cite{bucsoniu2010multi}. This extends the learned policies from exploring subsets of the state space to encompass the entire state space.

We consider policy distillation from \(k\) teacher networks to a student network. The first step in policy distillation involves collecting experiences by student agents in the source environments. Their experiences are stored separately as state-action pairs in each dataset \(\mathcal{D}_k=\{(s_i,a_i)\}_{i=1}^{N_k}\), where \(s_i \in \mathcal{S}\) are observations from the current state and \(a_i \in \mathcal{A}\) are the respective actions proposed by the policy \(\pi_\theta^{T_k}\), i.e., \(a_i=\pi_\theta^{T_k}(s_i)\). The student agent then learns from the \(k\) datasets sequentially. Instead of employing the KL divergence as the policy distillation loss \cite{hinton2015distilling}, we directly minimize the mean squared error between the deterministic actors \(\pi_\theta^{T_k}\) of the teachers and the deterministic actor of the student \(\pi_\theta^S\) for distillation:
\begin{align}
    \mathcal{L} _{MSE}(\pi_\theta^{T_k},\pi_\theta^S)=\frac{1}{|\mathcal{D}_k|} \sum_{i=1}^{|\mathcal{D}_k|}(\pi_\theta^{T_k}(s_i)-\pi_\theta^S(s_i))
\label{eq:2-7}
\end{align}
An important contribution of adopting policy distillation in geomagnetic navigation tasks is to obtain an agent-centric navigation model that overcomes the issue of model degradation in unexplored regions for autonomous mobile vehicles. The training of the student network is evenly interleaved between the datasets of multiple teacher networks, thereby more effectively combining the strategies of the teacher networks into a new network. This approach prevents the learned strategies from overfitting to the training environment and reduces the interference of geomagnetic field characteristics specific to a single environment on the navigation model.

\subsection{Reward shaping}
Previous DRL-based navigation studies generally design sparse reward for their learning agent, granting a reward only when the agent reaches the destination. Thus, the agent gains nothing from the environment until it reaches the navigation goal. This approach requires a large number of learning episodes and leads to significant sample inefficiency, especially in large-scale navigation spaces and complex state distributions as is in geomagnetic navigation.

A straightforward way to alleviate the sample efficiency problem is to add more information about the distance between the agent’s current state and goal to the reward design. The Earth's magnetic field, primarily generated by sources within the planet's outer core, exhibits a relatively uniform distribution over the Earth's surface. This uniformity means that geomagnetic parameters can provide consistent and reliable information, making them valuable for setting rewards in navigation tasks. In geomagnetic navigation, the closer the agent gets to the destination, the more similar the measured geomagnetic parameters at the current position to those of the destination. Therefore, using real-time geomagnetic data in the reward setting can integrate more information for the navigation, enhancing the agent's ability to navigate efficiently. Therefore, we use the objective function \(F(\mathbf{B}_j,j)\) defined in \Cref{eq:2-6}, where we compare the objective function at the current time step with the previous time step. This reward design results in that rewards are imposed on actions leading to the carrier moving towards the destination, while penalties are imposed on actions driving the carrier away from the destination. 

The agent interacting with the environment receives an extrinsic reward \(r_t^e\), which is calculated as follows:
\begin{align}
        r_t^e=\left\{\begin{matrix}
 r_{\text{goal} }, & F(\mathbf{B}, t)<\zeta\\
 \alpha(F(\mathbf{B},t) - F(\mathbf{B},t-1)), & F(\mathbf{B}, t)\ge \zeta
\end{matrix}\right.
        \label{eq:3-4}
\end{align}
where \(r_\text{goal}\) signifies the positive reward obtained by the vehicle successfully reaching the destination, and we use the weighted change in geomagnetic field parameters \(\alpha(F(\mathbf{B},t) - F(\mathbf{B},t-1)\) at different time steps as a reward.

However, simply using the extrinsic reward \(r_t^e\) for reward shaping may introduce additional problems. Although the agent explores and receives feedback in a multi-dimensional magnetic field space, the effects of its actions occur in the motion space of the autonomous mobile vehicle, this dimensional mismatch makes it challenging to capture the mapping relationship accurately. For instance, reaching a goal state might require first increasing a certain geomagnetic parameter before decreasing it, leading to additional local optima.

One solution to enhance learning efficiency is to allow agents to create appropriate intrinsic rewards, thereby making the rewards more abundant. \cite{xu2021magnetic} found that a key protein named cryptochrome 4 in the eyes of night-migratory birds, such as the European robin, is sensitive to magnetic fields. This sensitivity arises from its structure, which includes chains of amino acids that create magnetically sensitive radical pairs. When these birds are exposed to the Earth's magnetic field, these radical pairs react, providing directional information. For RL-based agents, utilizing magnetic information as an intrinsic reward signal can enable them to explore and navigate unexplored environments more effectively, avoiding pitfalls due to sparse feedback. The idea of using magnetic information as an intrinsic reward involves predicting the heading angle based on the current state \(s_t\) and the previous state \(s_{t-1}\), then calculating the error between the predicted heading angle and the actual heading angle from the action. This error serves as an intrinsic reward for the agent.

Inspired by the parallel approach method in missile guidance \cite{gong2022coordinated}, we introduce the gradient of geomagnetic parameters. At each time step, we calculate the heading angle that allows each geomagnetic parameter to converge towards the destination's geomagnetic values at the same rate, thus providing the agent with additional heading reference information. This can be expressed mathematically as follows:
\begin{align}
(B_i^{j+1}-B_i^j)\propto (B_i^T-B_i^j)
\label{eq:3-12}
\end{align}

By projecting vectors \((B_i^{j+1}-B_i^j)\) and \((B_i^T-B_i^j)\) onto the geographical coordinate system, the following results can be obtained:
\begin{align}
\frac{B_{i_1}^{j+1}-B_{i_1}^j}{B_{i_1}^T-B_{i_1}^j} =\frac{B_{i_2}^{j+1}-B_{i_2}^j}{B_{i_2}^T-B_{i_2}^j}
\label{eq:3-13}
\end{align}
where geomagnetic parameters at two adjacent time steps satisfy the following equation:
\begin{align}
\left\{\begin{matrix}B_{i_1}^{j+1}=B_{i_1}^j + g_{{i_1},x}^j\cdot \cos \theta_k + g_{{i_1},y}^j\cdot \sin \theta_k 
 \\B_{i_2}^{j+1}=B_{i_2}^j + g_{{i_2},x}^j\cdot \cos \theta_k + g_{{i_2},y}^j\cdot \sin \theta_k 
\end{matrix}\right.
\label{eq:b+14}
\end{align}
where \(g_{{i_1},x}^j\), \(g_{{i_1},y}^j\), \(g_{{i_2},x}^j\), \(g_{{i_2},y}^j\) are the gradients of \(B_{i_1}^j\) and \(B_{i_2}^j\). By substituting \Cref{eq:b+14} into \Cref{eq:3-13}, the predicted heading angle can be calculated as \Cref{eq:3-15}
\begin{align}
\lambda'_j = \arctan(\frac{(B_{i_1}^j-B_{i_1}^T) \cdot g_{i_2,x} - (B_{i_2}^j-B_{i_2}^T) \cdot g_{i_1,x}}{(B_{i_2}^j-B_{i_2}^T) \cdot g_{i_1,y} - (B_{i_1}^j-B_{i_1}^T) \cdot g_{i_2,y}})
\label{eq:3-15}
\end{align}
and the intrinsic reward \(r_t^i\) is set as follows:
\begin{align}
        r_t^i=\beta(\pi / 4 - |\lambda_j - \lambda'_j|)
    \label{eq:3-6}
\end{align}
the intrinsic reward is set to decrease the difference between the predicted heading angle and the action heading angle, penalizing discrepancies exceeding \(\pi/4\) radians, while rewarding actions that minimize the difference between them. 

The reward function is divided into two components: the agent interacting with the environment obtains the extrinsic reward \(r_t^e\), as shown in \cref{eq:3-4}, and the intrinsic reward \(r_t^i\) by navigation calculation based on the parallel approach method, as shown in Equation \cref{eq:3-6}. The agent is trained to optimize the policy to maximize the sum \(r_t=r_t^e+r_t^i\).

\section{Simulation and Evaluation}\label{Section:4}

In this section, we implement a custom geomagnetic navigation environment in OpenAI Gym and simulate the navigation of autonomous mobile vehicle in unexplored areas. To demonstrate the effectiveness of our method, we conduct comprehensive experiments and compare it with several baseline methods for geomagnetic navigation.

\subsection{Simulation Settings}

\subsubsection{Environments} We selected a large region spanning 45 degrees in longitude, from 90 degrees to 135 degrees, and 25 degrees in latitude, from 10 degrees to 35 degrees. and created the 2D grid matrix with OpenAI Gym to represent the geomagnetic field environment in this region. The geomagnetic field data is retrieved from the International Geomagnetic Reference Field Model (IGRF), a standard mathematical description of the large-scale structure of the Earth's main magnetic field and its secular variation. The IGRF was created by fitting parameters of a mathematical model to measure magnetic field data from surveys, observatories, and satellites across the globe \cite{thebault2015international}.

\begin{figure}[h]
    \centering
    \includegraphics[width=0.5\linewidth]{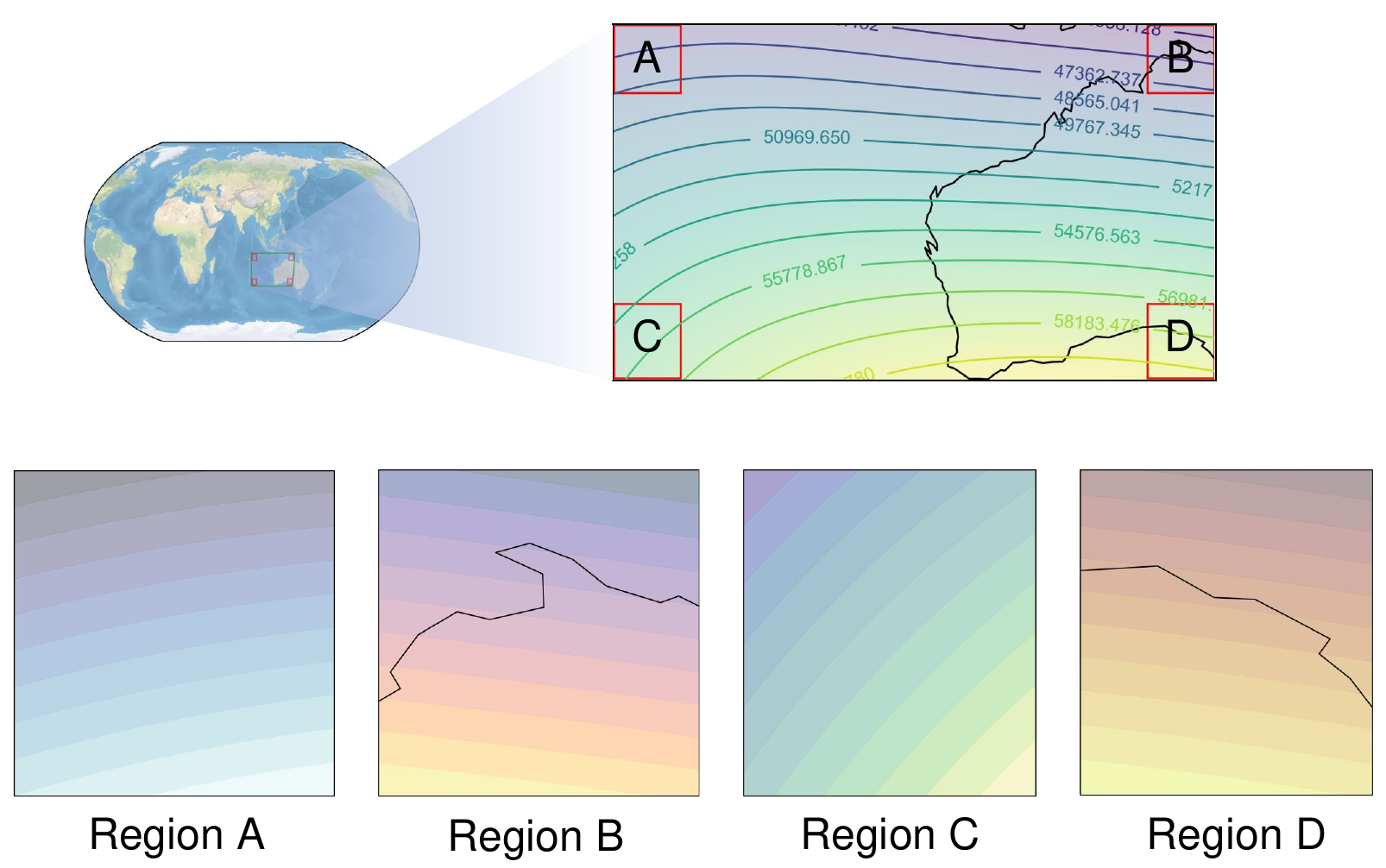}
    \caption{Total magnetic field intensity in the selected simulation region. The red boxes indicate Regions A, B, C, and D, which are used as training regions for the agent. The remaining region are used as unknown regions to validate the generalization of the navigation methods.}
    \label{fig:3-2}
\end{figure}

We perform simulations in this region designated for training and testing,  simulating scenarios such as vehicles completing long-distance navigation tasks in DNSS-denied environments using only values of the geomagnetic field measured by magnetometers. We selected subregions with a 5-degree span in both longitude and latitude at each of the four corners of the region, designated as \textbf{Region A}, \textbf{Region B}, \textbf{Region C}, and \textbf{Region D}, for training the agent. The remaining region serves as the unknown region for validating the performer of navigation algorithms. For simulation, we abstract the movement of autonomous mobile vehicles over large regions as the motion of a particle in a Cartesian coordinate system. The selected regions and their respective total field strengths are depicted in \Cref{fig:3-2}, emphasizing that each training region exhibits unique geomagnetic field distribution characteristics, which allows us to validate the generalization of various algorithms in different regions.

In all these scenarios, the vehicle is tasked with making navigation decisions based on the values of geomagnetic parameters of its current grid cell, accurately reaching the destination. During navigation, the heading angle \(\psi\) ranges from \(-2/\pi\) to \(2/\pi\), and the navigation distance \(L\) ranges from 0 to 50 \textit{km}. Therefore, at each time step, the agent predicts navigation actions based on the values of geomagnetic parameters of its current grid cell, guiding the vehicle to move \(L_t\) kilometers in the direction of the \(\psi _t\).

\subsubsection{Implementation} Our simulation experiments are conducted using a custom environment in OpenAI Gym. There are two types of training models: one is the teacher network, and the other is the student network. We divide the regions in the environment into multiple sub-regions, categorizing them as known and unknown sub-regions. Each known sub-region is assigned a TD3 agent to train an independent teacher network.

The TD3 agent's hyperparameters were optimized using the Optuna tool to enhance performance \cite{akiba2019optuna}. The training process involves 600,000 timesteps to ensure adequate learning. We set the discount factor \(\gamma\) to 0.995 to prioritize long-term rewards. The agent's experience buffer size is 50,000, allowing it to store a substantial amount of past experiences for learning. We introduce 'normal' noise with a standard deviation of 0.1 to encourage exploration and prevent overfitting. During training, gradient updates are performed at every step with a batch size of 256, ensuring efficient learning and stability. The policy network is architected with multi-layer perceptrons (MLPs) consisting of layers with 400 and 300 units, providing a robust structure for approximating complex functions. The training frequency is set to update the network at each timestep, and the learning rate is 0.0003, balancing the speed and stability of learning.

We systematically analyze the training process of the teacher networks in their respective sub-regions to determine the average reward and average navigation time at convergence. The student network adopts the same configuration as the teacher network and is trained using a dataset composed of samples from the experience replay buffers of all student networks.

During testing, we primarily assess the navigation performance of both the teacher and student networks in unknown sub-regions, exploring and comparing their adaptability to unexplored areas. It’s worth noting that the simulations were executed on a platform equipped with an RTX 3080 Ti GPU, i7-9700K CPU, and 32 GB of RAM.

\subsubsection{Ablation methods and baselines} We compare our method with some ablation methods and state-of-the-art methods. We consider three ablation methods, including \textbf{TD3-SR}, \textbf{TD3-ER}, and \textbf{TD3-ST}. We also consider three baseline methods, including \textbf{PSO}, \textbf{ACA}, \textbf{DE}, and \textbf{GA}, as baselines for assessing performance and conducting comparative analyses, with the aim of substantiating the efficacy of the reward shaping and policy distillation mechanism. The distinctive attributes of each model are delineated as follows.

\textbf{TD3-SR} (\textbf{TD3} with \textbf{S}parse \textbf{R}ewards) and \textbf{TD3-ER}(\textbf{TD3} with \textbf{E}xtrinsic \textbf{R}ewards) are two ablation methods based on the TD3 framework. These methods are primarily used to validate the effectiveness of the improved reward reshaping method proposed in this paper. The reward functions for these two methods are set as shown in \Cref{eq:4-1} and \Cref{eq:4-2}:

\begin{align}
        r_\text{TD3-SR}=r_{\text{goal}}
        \label{eq:4-1}
\end{align}
\begin{align}
        r_\text{TD3-ER}=r_{\text{goal}}+\alpha(F(\mathbf{B},t) - F(\mathbf{B},t-1))
        \label{eq:4-2}
\end{align}
The reward function for \textbf{TD3-SR} grants a reward only when the vehicle reaches its destination, representing a typical sparse reward setting. The reward function for \textbf{TD3-ER} includes an additional weighted difference in the normalized objective function as shown in \Cref{eq:2-6} between adjacent time steps, reflecting how close the values of geomagnetic parameters at the current position are to those at the destination, but it does not utilize the intrinsic reward as shown in \Cref{eq:3-6} designed in this paper. Comparing these two methods with \textbf{TD3-ST} aims to validate the improvement in exploration efficiency for long-distance geomagnetic navigation achieved by the reward shaping method proposed in this article. 

\begin{figure*}[htb]
\centering
\subfloat[]{\includegraphics[width=3.2in]{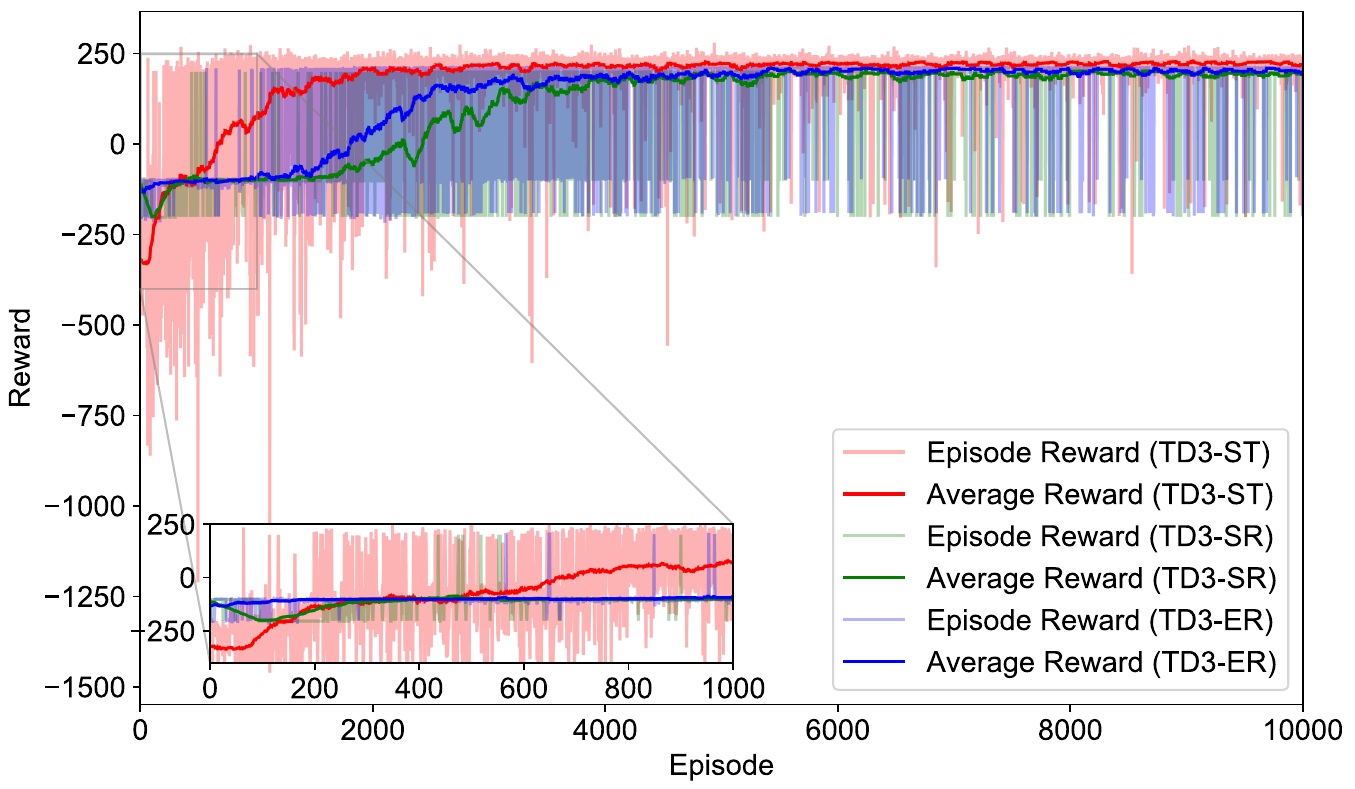}%
\label{fig_first_case}}
\hfil
\subfloat[]{\includegraphics[width=3.2in]{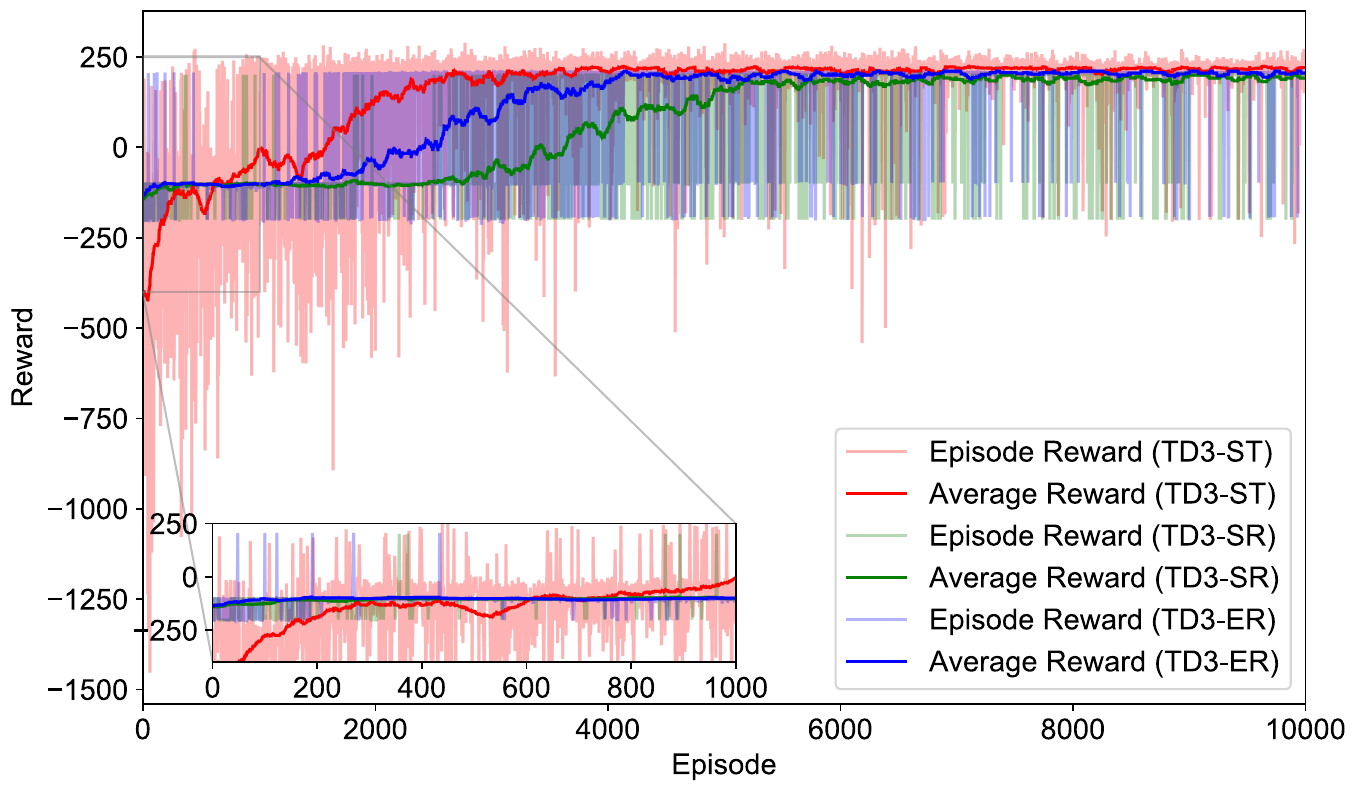}%
\label{fig_second_case}}
\vfil
\subfloat[]{\includegraphics[width=3.2in]{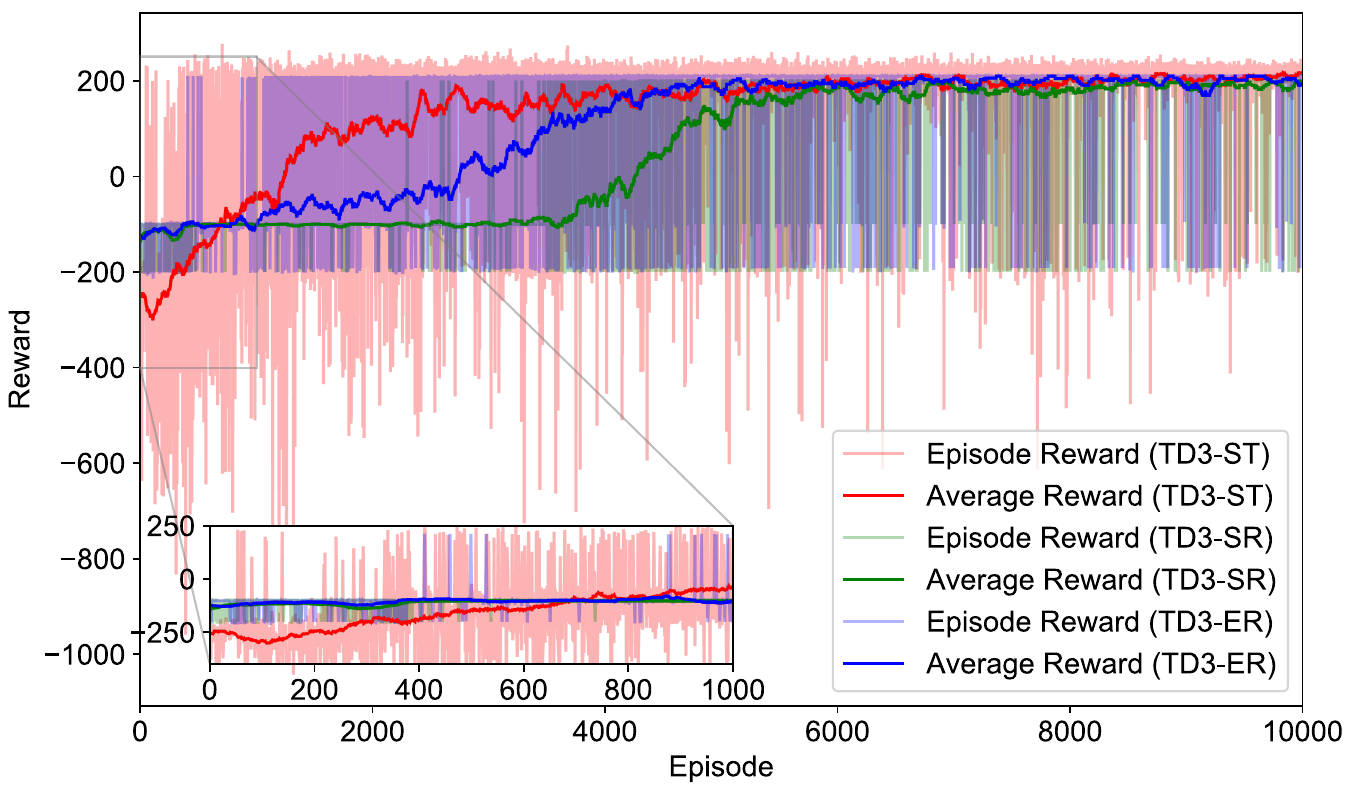}%
\label{fig_third_case}}
\hfil
\subfloat[]{\includegraphics[width=3.2in]{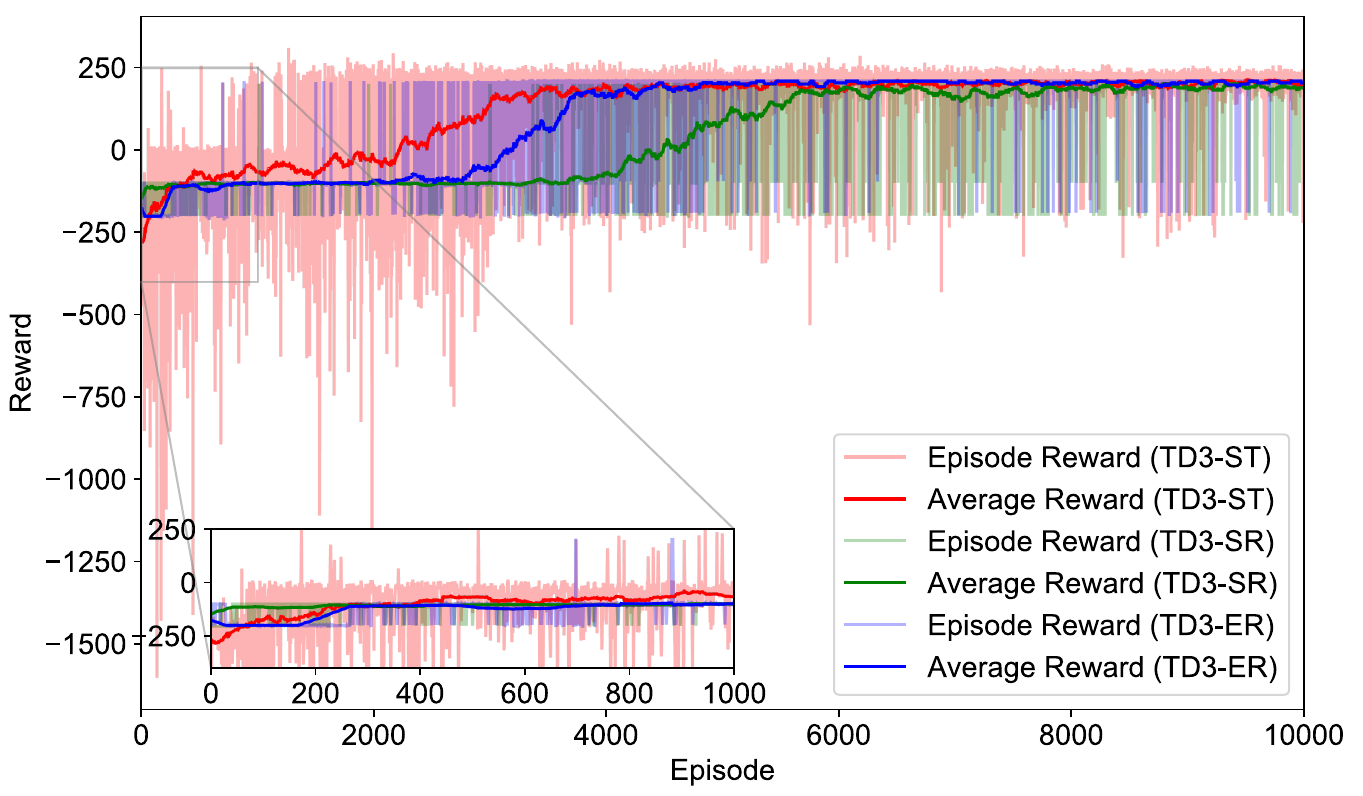}%
\label{fig_fourth_case}}
\caption{Reward curves for the TD3-ST algorithm and two ablation methods, TD3-SR and TD3-ER, trained in four regions: (a) Region A: Longitude 90 to 95, Latitude -15 to -10, (b) Region B: Longitude 130 to 135, Latitude -15 to -10, (c) Region C: Longitude 90 to 95, Latitude -35 to -30, and (d) Region D: Longitude 130 to 135, Latitude -35 to -30. Each subfigure displays the convergence behavior of the algorithms, with both raw and averaged episode rewards.}
\label{fig:5.1}
\end{figure*}

\textbf{TD3-ST} (\textbf{TD3} with \textbf{S}eparate \textbf{T}raining) is the model that uses the reward function as follows:
\begin{align}
        r_\text{TD3-ST}=\underbrace{r_{\text{goal}}+\alpha(F(\mathbf{B},t) - F(\mathbf{B},t-1))}_{r_{\text{extrinsic}}}+r_{\text{intrinsic}}
        \label{eq:4-3}
\end{align}
\textbf{TD3-ST} serves as the teacher network for the proposed \textbf{TD3-STEPD}. After being trained in sub-regions, it is directly used for navigation testing in unknown regions without ensemble policy distillation. This is done to verify the effectiveness of ensemble policy distillation in enhancing performance across unexplored scenarios.

\textbf{PSO} (\textbf{P}article \textbf{S}warm \textbf{O}ptimization) \cite{kennedy1995particle}, \textbf{AFSA} (\textbf{A}rtificial \textbf{F}ish \textbf{S}warm \textbf{A}lgorithm) \cite{yazdani2011fuzzy}, \textbf{DE} (\textbf{D}ifferential \textbf{E}volution) \cite{storn1997differential}, and \textbf{GA} (\textbf{G}enetic \textbf{A}lgorithm) \cite{holland1975adaptation} are four baseline methods belonging to metaheuristic algorithms and are currently the mainstream methods adopted in geomagnetic navigation research.

These three methods share common characteristics as metaheuristic algorithms. Metaheuristic algorithms are designed to find solutions for complex optimization problems within a reasonable amount of time, without guaranteeing the absolute optimal solution. They are characterized by their ability to explore and exploit the search space efficiently, handle large and complex problem domains, and provide robust solutions through iterative improvement processes. 
We use these four Metaheuristic methods as baseline methods to explore their efficacy in varying navigation scenarios.

\subsubsection{Evaluation metrics} When evaluating the methods, we report the experimental results in terms of Success Rate (SR) \cite{sang2022novel}, Success Weighted by Path Length (SPL) \cite{batra2020objectnav}, Mean Absolute Error of Heading Deviation \cite{sabet2017low}, Root Mean Square Error of Heading Deviation \cite{sabet2017low}, Total Navigation Time (TNT) \cite{zhang2022ipaprec}, and Navigation Error (NE) \cite{lin2023towards}. We first report the reward curves to indicate the impact on exploration efficiency, then compared the navigation performance of various ablation methods based on TD3 with the proposed TD3-STEPD in the training regions, reporting SR and SPL to reflect the navigation efficiency and success rate. Then we compare the navigation performances in explored region, contrasting the generalization abilities of different methods. 
Furthermore, we compare the navigation trajectory through box plots of Absolute Mean Error and Root Mean Square Error of Heading Deviation, as well as NE and TNT. Finally, we visualize navigation trajectories, comparing the trajectories by the proposed DRL approach with evolutionary geomagnetic navigation methods.

\subsection{Simulation Result}
We train four teacher networks, the reward curves are drawn as shown in \Cref{fig:5.1}, we can get the following observations:

\begin{itemize}
    \item The reward of 200 indicates the vehicle reaches its destination, which is significantly higher than other rewards. The average reward reaches 200 in all four regions no later than 6000 training episodes. This indicates that the proposed method and the two ablation methods can stabilize and complete the random navigation tasks within their respective training regions after a certain number of training episodes. The rapid increase in rewards suggests effective exploration, while the subsequent stabilization at higher reward values reflects efficient exploitation of the learned policies.
    \item Different regions influence the agent's convergence speed. For example, the reward curves in Region C for all three methods rise more slowly than in Region D. This reflects how variations in the geomagnetic field environment affect navigation performance. Such differences underscore the importance of environmental conditions in training effectiveness and the adaptability of navigation algorithms.
    \item By analyzing the average reward curves, it is evident that the TD3-ST consistently outperforms the two ablation methods across all four regions. This demonstrates that the proposed method enables the agent to learn effective navigation strategies more quickly.
    \item From reward curves within the first 1000 training episodes, the proposed TD3-ST reaches its destination for the first time in episodes 54, 65, 174, and 13 in the four regions respectively. In comparison, TD3-SR achieves this in episodes 2298, 437, 1012, and 357, while TD3-ER does so in episodes 412, 566, 698, and 48. Considering the high cost of exploration in long-distance navigation, our proposed method’s additional intrinsic rewards significantly improve exploration efficiency. This allows the agent to quickly gather positive samples, thereby accelerating algorithm convergence.
\end{itemize}

\subsection{Numerical Evaluation on Cross-Region Navigation}

\begin{table*}[htb]
\caption{Evaluation of Proposed and Ablation Methods in the Training Regions, Focusing on SR and SPL. The Horizontal Axis Represents the Metrics of Algorithms Tested in Their Respective Training Regions.}
\centering
\resizebox{0.65\textwidth}{!}{%
\begin{tabular}{ccccccccc}
\hline

\multirow{2}{*}{\textbf{Test environment}}& \multicolumn{2}{c}{\textbf{TD3-SR}} & \multicolumn{2}{c}{\textbf{TD3-ER}} & \multicolumn{2}{c}{\textbf{TD3-ST}} & \multicolumn{2}{c}{\textbf{TD3-STEPD}} \\
 & \textbf{SR \textperthousand} & \textbf{SPL \textperthousand}& \textbf{SR \textperthousand} & \textbf{SPL \textperthousand}& \textbf{SR \textperthousand} & \textbf{SPL \textperthousand}& \textbf{SR \textperthousand} & \textbf{SPL \textperthousand}\\
\hline
Region A & 985 & 906.03 & 989 & \textbf{948.04} & \textbf{995} & 888.62 & 989 & 880.98 \\
Region B & 990 & \textbf{928.62} & 994 & 894.48 & 995 & 919.69 & \textbf{997} & 920.02 \\
Region C & 984 & 916.78 & 996 & 925.74 & \textbf{999} & \textbf{953.57} & 997 & 949.37 \\
Region D & 995& 946.6 & 997 & 946.96 & 996 & 964.74 & \textbf{998} & \textbf{965.91} \\
\hline
\end{tabular}%
}
\label{tab:3-1}
\end{table*}

\begin{table*}[htb]
\centering
\caption{Evaluation of Proposed and Ablation Methods in the Unknown Region, Focusing on SR and SPL. The Horizontal Axis Represents the Metrics of Models Tested in Unknown Environments, which were Trained in Specific Training Environments. Since the TD3-STEPD model is Obtained Through Policy Distillation, There is Only One Set of Results Available.}
\label{tab:3-2}
\resizebox{0.98\textwidth}{!}{%
\begin{tabular}{|c|ll|ll|ll|ll|ll|ll|ll|ll|}
\hline
\multirow{2}{*}{\tabincell{c}{\textbf{Training}\\ \textbf{environment}}} &
  \multicolumn{2}{c|}{\textbf{TD3-SR}} &
  \multicolumn{2}{c|}{\textbf{TD3-ER}} &
  \multicolumn{2}{c|}{\textbf{TD3-ST}} &
  \multicolumn{2}{c|}{\textbf{TD3-STEPD}} &
  \multicolumn{2}{c|}{\textbf{AFSA}} &
  \multicolumn{2}{c|}{\textbf{DE}} &
  \multicolumn{2}{c|}{\textbf{GA}} &
  \multicolumn{2}{c|}{\textbf{PSO}} \\
 &
  \textbf{SR \textperthousand} &
  \textbf{SPL \textperthousand} &
  \textbf{SR \textperthousand} &
  \textbf{SPL \textperthousand} &
  \textbf{SR \textperthousand} &
  \textbf{SPL \textperthousand} &
  \textbf{SR \textperthousand} &
  \textbf{SPL \textperthousand} &
  \textbf{SR \textperthousand} &
  \textbf{SPL \textperthousand} &
  \textbf{SR \textperthousand} &
  \textbf{SPL \textperthousand} &
  \textbf{SR \textperthousand} &
  \textbf{SPL \textperthousand} &
  \textbf{SR \textperthousand} &
  \textbf{SPL \textperthousand} \\ \hline
Region A &
  \multicolumn{1}{l}{2} &
  1.33 &
  \multicolumn{1}{l}{447} &
  341.72 &
  \multicolumn{1}{l}{475} &
  399.85 &
  \multicolumn{1}{l}{\multirow{4}{*}{\textbf{941}}} &
  \multirow{4}{*}{\textbf{823.91}} &
  \multirow{4}{*}{608} &
  \multirow{4}{*}{447.95} &
  \multirow{4}{*}{395} &
  \multirow{4}{*}{355.28} &
  \multirow{4}{*}{460} &
  \multirow{4}{*}{392.96} &
  \multirow{4}{*}{326} &
  \multirow{4}{*}{284.87} \\
Region B &
  \multicolumn{1}{l}{8} &
  4.33 &
  \multicolumn{1}{l}{490} &
  383.66 &
  \multicolumn{1}{l}{539} &
  430.16 &
  \multicolumn{1}{l}{} &
   &
   &
   &
   &
   &
   &
   &
   &
   \\
Region C &
  \multicolumn{1}{l}{14} &
  12.58 &
  \multicolumn{1}{l}{379} &
  302.27 &
  \multicolumn{1}{l}{611} &
  443.35 &
  \multicolumn{1}{l}{} &
   &
   &
   &
   &
   &
   &
   &
   &
   \\
Region D &
  \multicolumn{1}{l}{14} &
  9.91 &
  \multicolumn{1}{l}{625} &
  482.89 &
  \multicolumn{1}{l}{699} &
  573.27 &
  \multicolumn{1}{l}{} &
   &
   &
   &
   &
   &
   &
   &
   &
   \\ \hline
\end{tabular}%
}
\end{table*}

We increased the training iterations of the agents to 30,000 in their respective environments. The trained TD3-SR served as the teacher network and trained the student network TD3-STEPD by policy distillation. We first randomly set up 1000 navigation tasks in each training region to comprehensively evaluate the performance of different ablation methods in executing these local navigation tasks and compare them with the proposed TD3-STEPD. These tasks had distances between 30 km and 50 km, ensuring consistency across different methods within the same region. Each episode had a maximum of 250 steps, with episodes deemed failures if they exceeded this limit. Evaluation metrics are shown in \Cref{tab:3-1}, with a focus on  \textbf{SR} and \textbf{SPL}. Furthermore, we validated the effectiveness of our proposed TD3-SREPD in unexplored regions by randomly setting up 1000 tasks outside the training regions. The results of \textbf{SR} and \textbf{SPL} are shown in \Cref{tab:3-2}, by comparing these results, we can make the following observations:

\begin{figure}[htb]
\centering
\subfloat[]{\includegraphics[width=0.4\linewidth]{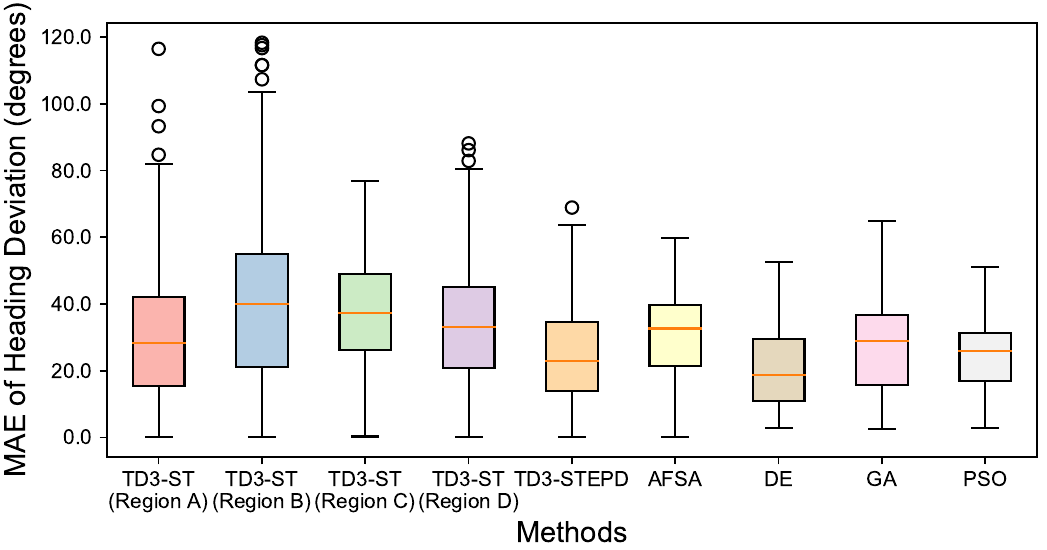}%
\label{fig:sub1}}
\subfloat[]{\includegraphics[width=0.4\linewidth]{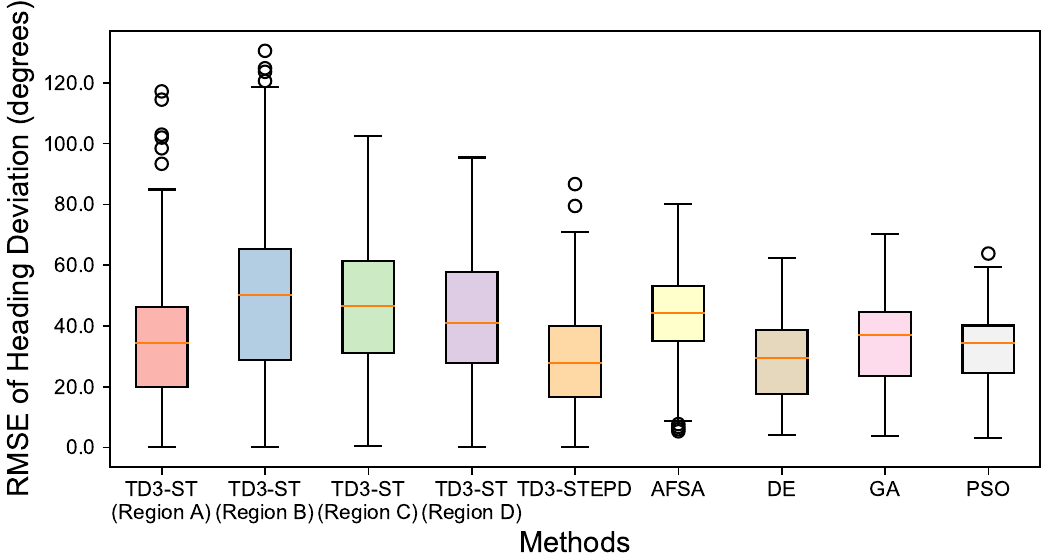}%
\label{fig:sub2}}
\vfill
\subfloat[]{\includegraphics[width=0.4\linewidth]{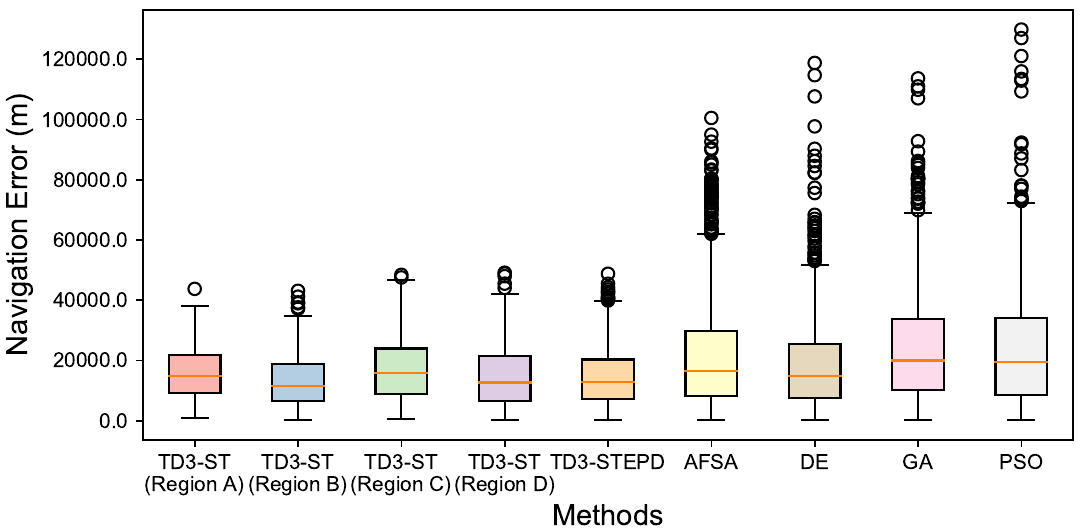}%
\label{fig:sub3}}
\subfloat[]{\includegraphics[width=0.4\linewidth]{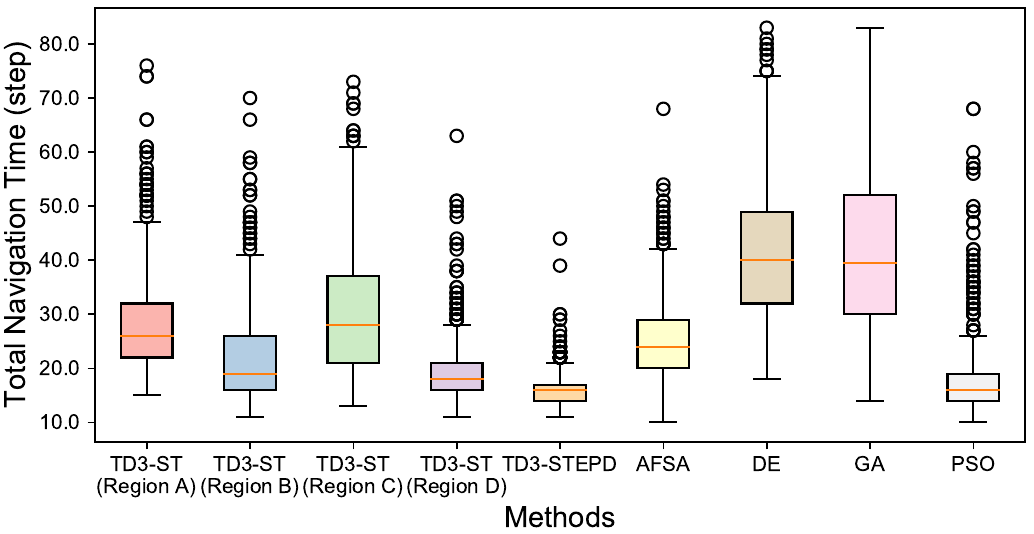}%
\label{fig:sub4}}
\caption{Statistical results for the comparison of four evaluation metrics between basic TD3-ST and proposed TD3-STEPD algorithms in 1000 long-distance navigation tasks within unknown regions. Each subplot corresponds to a specific evaluation metric: (a) Absolute Mean Error of Heading Deviation, (b) Root Mean Square Error of Heading Deviation, (c) Navigation Error and (d) Total Navigation Time. In each subplot, the central mark indicates the median, and the bottom and top edges of the box indicate the 25th and 75th percentiles, respectively. The whiskers extend to show the range of the data excluding outliers, which are represented by individual points beyond the whiskers.}
\label{fig:5.2}
\end{figure}

\begin{itemize}
    \item As can be seen from the \Cref{tab:3-1}, all the ablation algorithms, including the sparse reward-based TD3-SR, dense reward-based TD3-ER, and the more complex intrinsic reward-based TD3-ST, effectively utilized geomagnetic field parameters to achieve navigation tasks in their respective training regions by extensive training, the policy distillation-based TD3-STEPD also demonstrated successful navigation across all training regions. These algorithms achieved a navigation success rate of over 98\%, highlighting the feasibility of using deep reinforcement learning for geomagnetic navigation. Among the algorithms, TD3-ST and TD3-STEPD generally outperform the others. TD3-ST achieved the highest navigation success rates in \textbf{Region A} and \textbf{Region C}, while TD3-STEPD excelled in other regions. Notably, despite the high effectiveness of TD3-STEPD across various training regions, it did not consistently outperform its ablation counterpart, TD3-ST. This discrepancy can be attributed to the inherent complexity of policy distillation, where learning from multiple tasks in different regions introduces mapping changes. The transfer of navigation models from environment-centric to agent-centric behavior representations may have resulted in the loss of some environmental feature details, thereby impacting task execution.
    \item As indicated in \Cref{tab:3-2}, the performance of the algorithms in unknown environments exhibits notable dissimilarities compared to their performance in training environments. Four methods suffer from varying degrees of performance degradation, however, our proposed TD3-STEPD still achieves a navigation success rate of over 94\%, while the navigation success rates of the other three algorithms are generally below 70\%. This observation confirms that our method demonstrates outstanding generalization capabilities when applied to unexplored regions with geomagnetic field environments differing from those of the training regions. As a result, it circumvents the risk of overfitting the navigation model to the training environment, ensuring both flexibility and scalability. The excellent generalization ability of TD3-STEPD may be due to the use of ensemble policy distillation. It leverages high-level and intermediate-level knowledge from multiple teacher networks, enabling the student network to capture a broader range of features and patterns and helping create a more robust and flexible model that performs well across diverse and unseen environments.
\end{itemize}

To evaluate the performance of the methods in navigation tasks within unknown regions more accurately, we further analyzed the trajectories of the method before and after improvement using policy distillation in 1000 long-distance navigation tasks. Evaluation metrics included the Absolute Mean Error and Root Mean Square Error of Heading Deviation, reflecting the straightness of trajectories, Navigation Error reflecting navigation accuracy, and Total Navigation Time reflecting navigation efficiency, as shown in \Cref{fig:5.2}. We draw the following observations from the figure:

\begin{itemize}
    \item The proposed TD3-STEPD demonstrates lower MAE and RMSE of heading deviation compared to the individually trained TD3-ST across the four regions. This indicates that multi-teacher policy distillation aids in developing robust navigation policies by incorporating diverse experiences and mitigating overfitting to specific training environments. Consequently, this allows the autonomous mobile vehicle to effectively determine the direction towards the destination in unknown regions by analyzing the variations in geomagnetic field parameters during its movement.
    \item The navigation error for TD3-ST agents varies across different regions, with the highest error observed in \textbf{Region A} and the best performance in \textbf{Region B}. The proposed TD3-STEPD demonstrates navigation error that fall between the four TD3-ST agents. This suggests that TD3-STEPD adopts navigation strategies similar to those of its TD3-ST teachers when approaching the destination.
    \item The proposed TD3-STEPD significantly reduces the total navigation time in unknown regions. This improvement can be attributed to the enhanced generalization capability provided by policy distillation from multiple teachers. Multi-teacher policy distillation accelerates learning and increases efficiency by synthesizing diverse strategies into a unified policy. This allows the autonomous mobile vehicle to navigate more efficiently and quickly adapting to new environments.
\end{itemize}

\begin{figure*}[!t]
\centering
\subfloat[]{\includegraphics[width=2.3in]{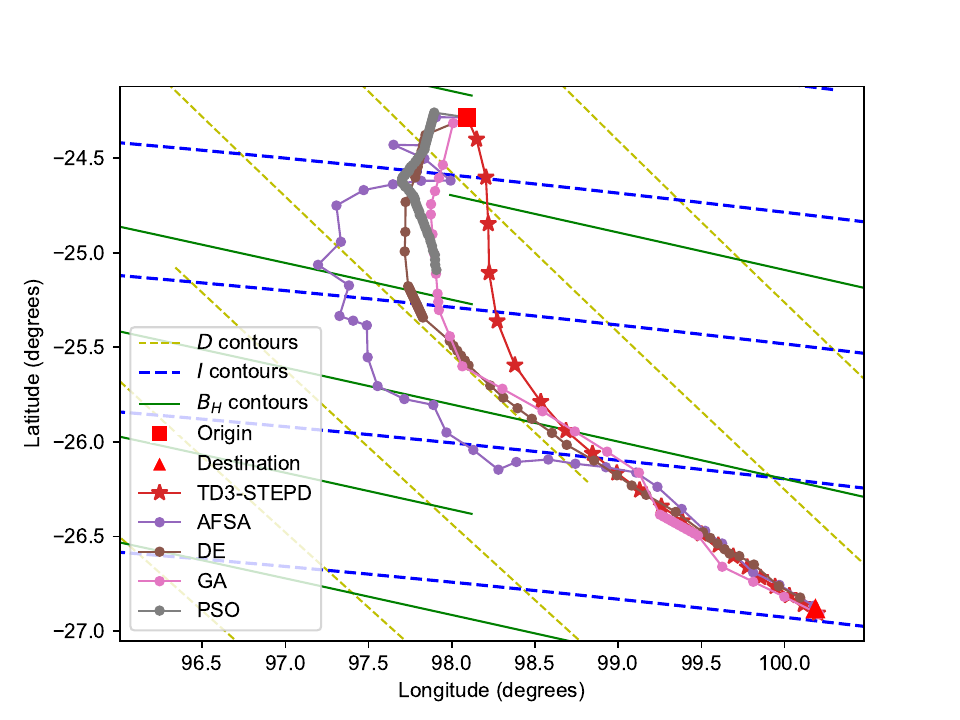}%
\label{zhi_hui}}
\hfil
\subfloat[]{\includegraphics[width=2.3in]{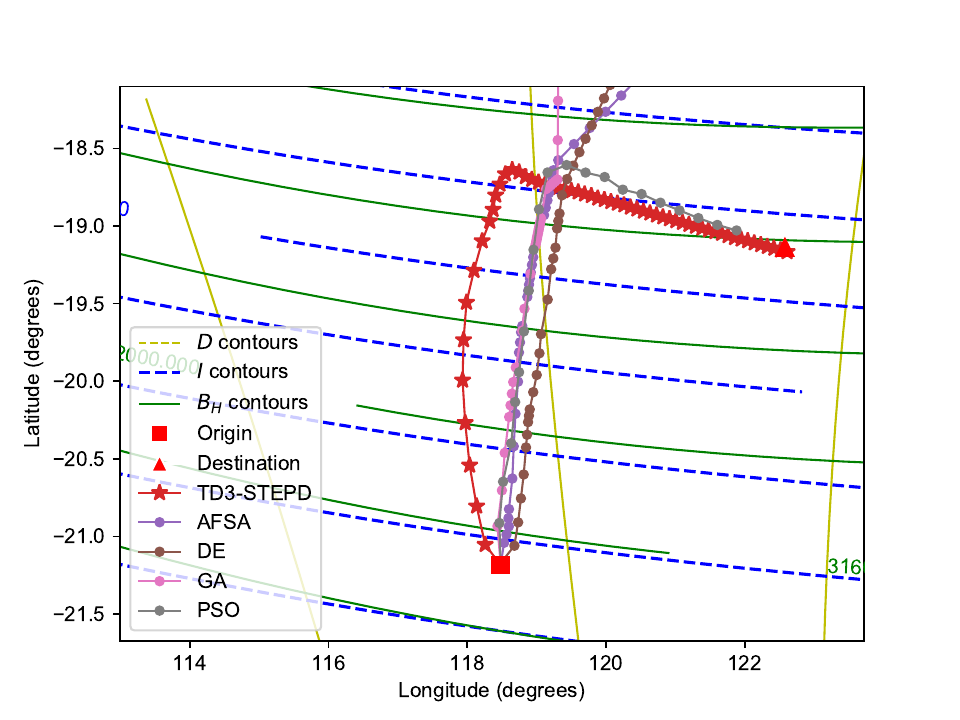}%
\label{zhe1}}
\hfil
\subfloat[]{\includegraphics[width=2.3in]{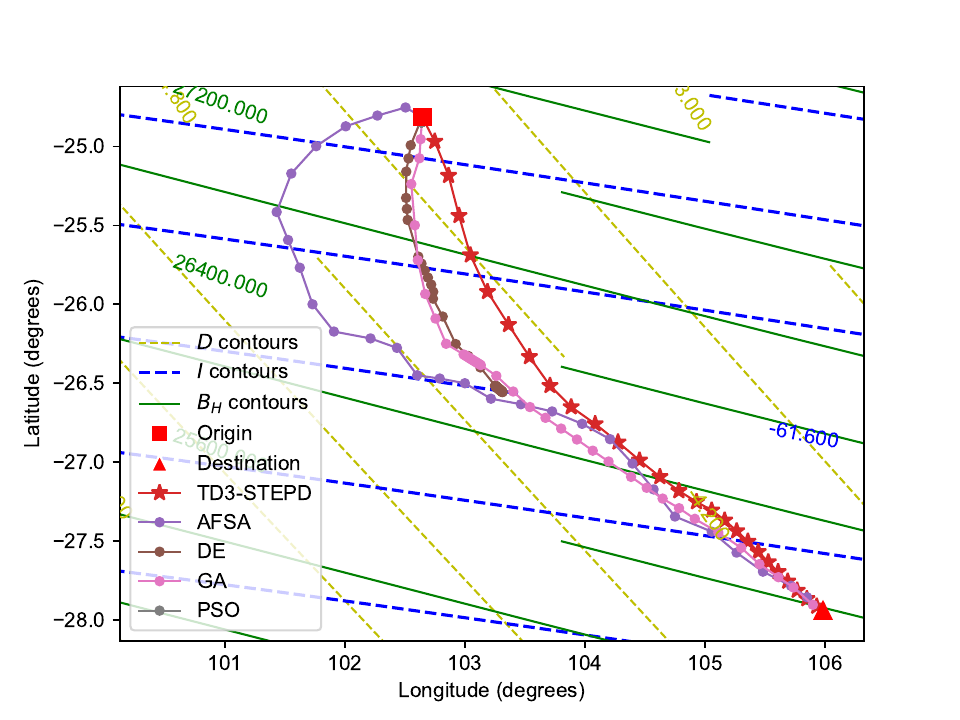}%
\label{zhi2}}
\vfil
\subfloat[]{\includegraphics[width=2.3in]{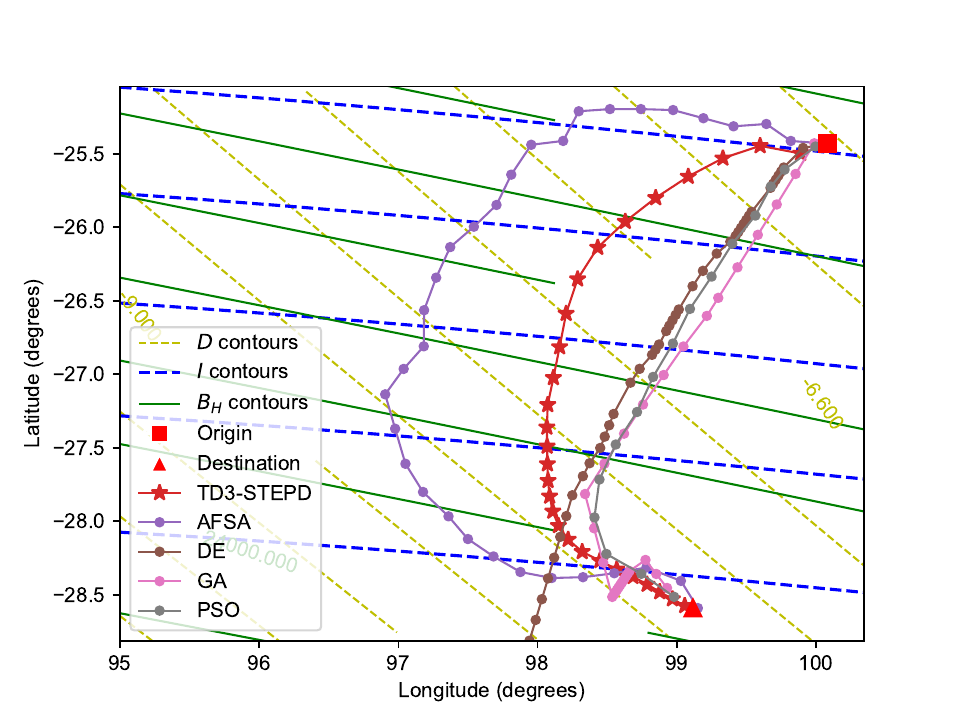}%
\label{zhe3}}
\hfil
\subfloat[]{\includegraphics[width=2.3in]{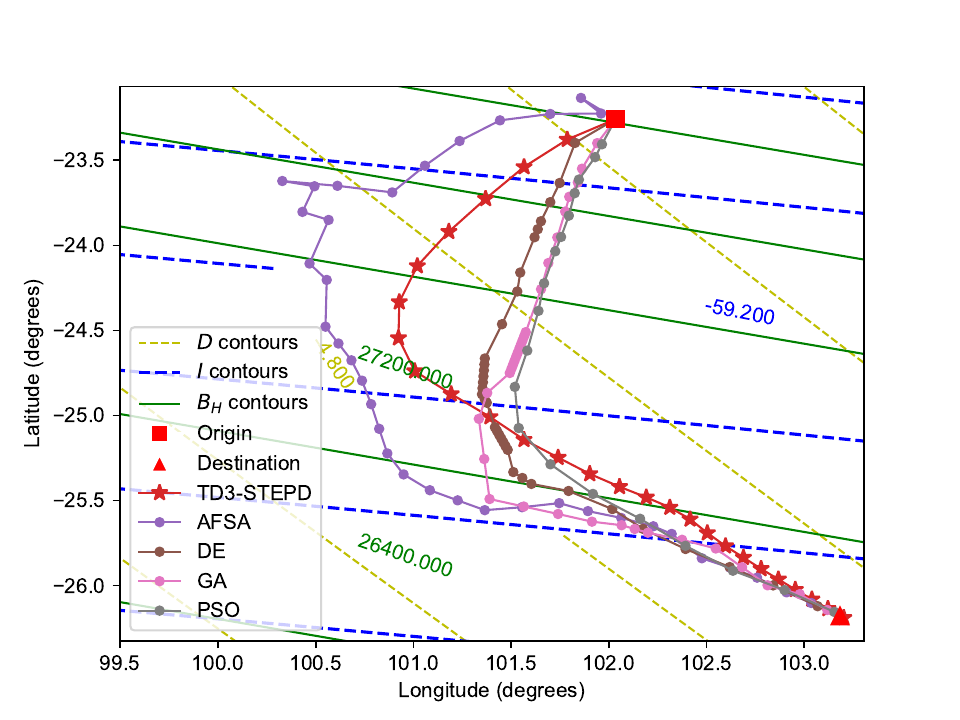}%
\label{shun1}}
\hfil
\subfloat[]{\includegraphics[width=2.3in]{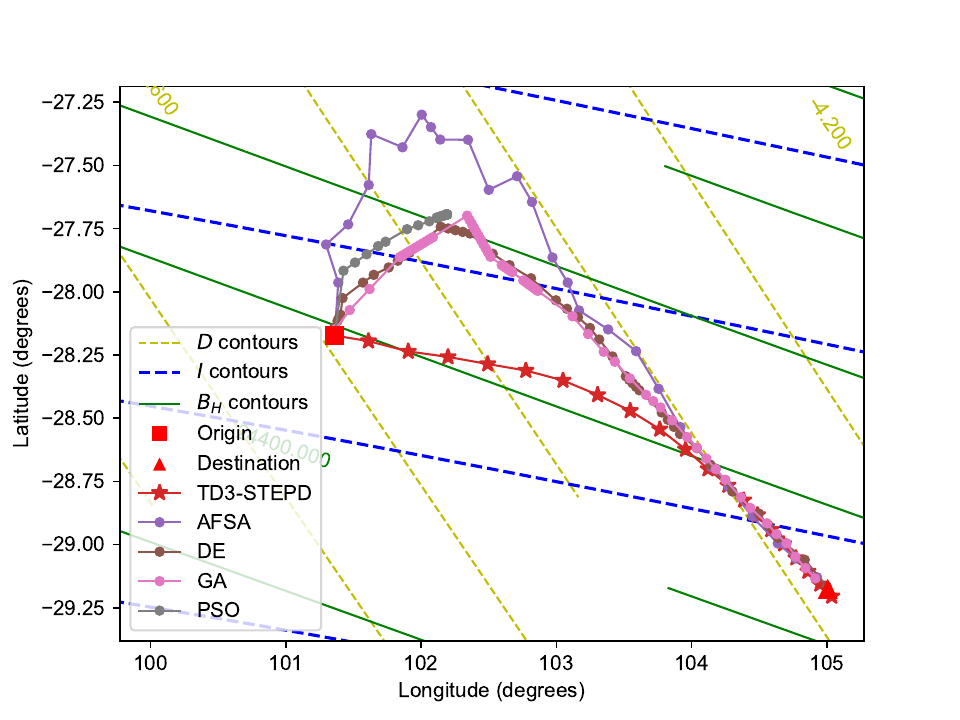}%
\label{zhi3}}
\vfil
\subfloat[]{\includegraphics[width=2.3in]{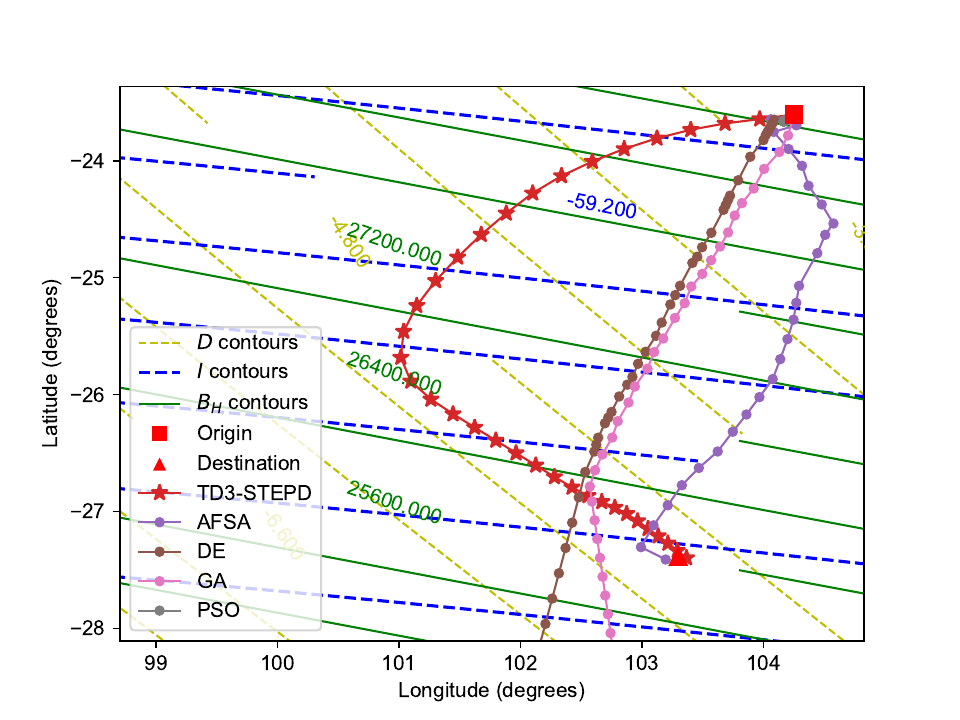}%
\label{shun_nihe}}
\hfil
\subfloat[]{\includegraphics[width=2.3in]{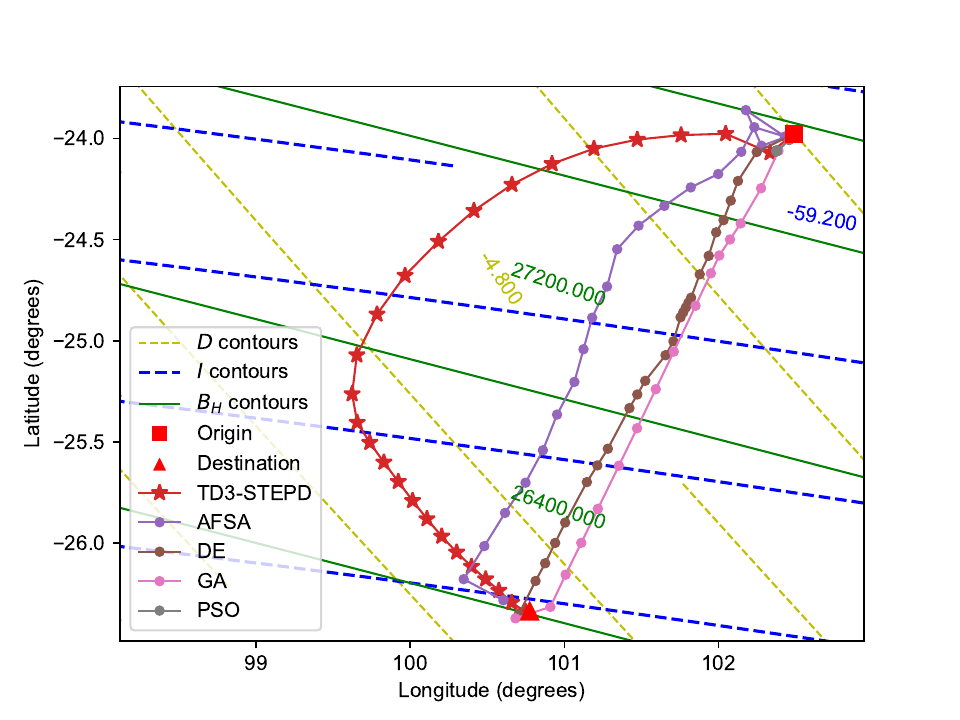}%
\label{fan}}
\hfil
\subfloat[]{\includegraphics[width=2.3in]{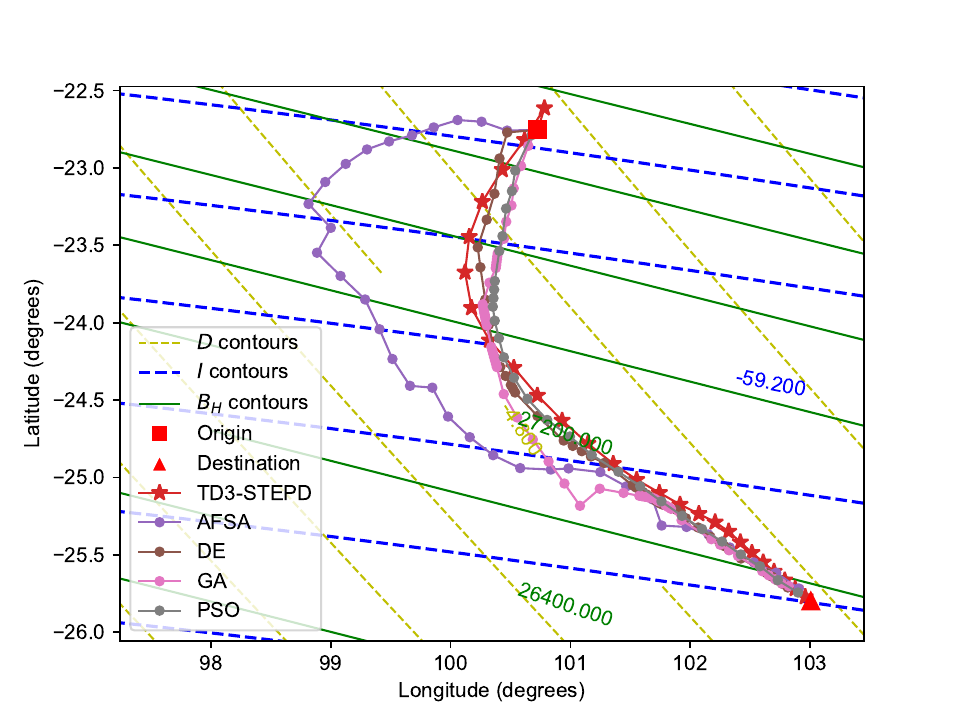}%
\label{shun2}}
\vfil
\caption{Nine distinct navigation tasks for the proposed TD3-STEPD and four baseline methods. The red, purple, brown, pink, and gray lines represent the trajectories taken by the agent of TD3-STEPD and the four baseline methods. The red rectangle and triangle markers denote the origin position and the destination of the navigation task, respectively.}
\label{fig:5.3}
\end{figure*}

\subsection{Visualization Evaluations on Navigation Trajectories}

We set up multiple navigation tasks to evaluate and compare our proposed reinforcement learning-based TD3-STEPD with baseline methods from previous studies based on metaheuristic algorithms, these methods are AFSA, DE, GA, and PSO, respectively. The fitness function of these methods is defined in \Cref{eq:2-6}. The visual trajectories are presented in \Cref{fig:5.3},  and this function is employed as the dense reward function in the proposed TD3-STEPD. By observing the visual results,  we can draw several insights:

\begin{itemize}
    \item The navigation trajectories generated by the proposed TD3-STEPD are significantly smoother compared to the four baseline methods. This smoothness can be attributed to the learning-based mechanism that derives strategies from extensive experience. Through the learning process, the TD3-STEPD can learn a more optimal and stable navigation strategy, reducing the randomness and instability often associated with metaheuristic algorithms that rely on random exploration. The TD3-STEPD also exhibits the ability to gradually decrease the navigation distance at each time step as it approaches the destination, which is not observed in the baseline methods. The ability to make finer adjustments to the navigation distance as the vehicle nears the destination contributes to higher navigation accuracy, a trait consistent with the results observed in \Cref{fig:sub3}.
    \item When considering the efficiency of navigation, the proposed TD3-STEPD does not always hold an advantage. For instance, as shown in \Cref{fan}, the navigation trajectory of the four baseline methods is noticeably shorter than that of TD3-STEPD. Conversely, in \Cref{zhi2}, TD3-STEPD outperforms the baseline methods by achieving a shorter trajectory. In other results, the length of TD3-STEPD's navigation trajectory typically falls between those of the baseline methods. This variability indicates that the navigation efficiency of TD3-STEPD is influenced by the geomagnetic field characteristics of the specific region.
    \item The proposed TD3-STEPD demonstrates the most stable performance, which is corroborated by the success rate in \Cref{tab:3-2}. For example, in \Cref{zhe1} and \Cref{shun_nihe}, The baseline methods often fail to balance the convergence of all three geomagnetic parameters, sometimes prematurely fitting to the data when only one or two parameters have nearly converged, resulting in an inability to accurately adjust the heading angle. The TD3-STEPD method, however, can maintain consistent and stable navigation by effectively integrating navigation policies from multiple agents. It learns agent-centric behavior representations to adapt to different magnetic field environments, thus avoiding early convergence and ensuring more reliable navigation.
\end{itemize}

\section{Conclusion}\label{Section:5}

This work investigates the generalizability of a trained geomagnetic navigation model, and proposes a novel navigation solution based on deep reinforcement learning and multi-agent policy distillation. 
We design the reward in the learning agent with mixed reward to improve exploration efficiency and avoid ineffective exploration in large-scale regions. We adopt policy distillation to extract agent-centric behavior representations from navigation strategies trained across multiple regions. 
By integrating navigation strategies learned under varying geomagnetic environments, the distilled strategy effectively bridges the discrepancies in geomagnetic fields, thus adapting more robustly to navigation in unknown regions. The simulation results demonstrate that 
the proposed approach outperforms state-of-the-art evolutionary algorithm-based navigation strategies with higher success rates, efficiency, and accuracy, and the distilled strategy provides robust generalization capabilities when the learned model is applied to unexplored regions. 

We suggest investigations on the impact of geomagnetic anomalies on the navigation in the future work. There are dynamics in the geomagnetic field that can affect the navigation to different levels, \textit{e.g.}, by daily geomagnetic fluctuations, geomagnetic abnormalities, and geomagnetic storms. It is challenging to model those uncertainties precisely, while robust solutions for the geomagnetic navigation is essential especially in long-range missions where such anomalies are inevitable. An impact assessment for the anomalies, therefore, can provide reasonable evaluation on whether the geomagnetic navigation works under different levels of uncertainties.

\bibliographystyle{IEEEtran}
\bibliography{Reference}

\end{document}